\documentclass[10pt,twocolumn,letterpaper]{article}

\usepackage{cvpr}              % To produce the CAMERA-READY version
\usepackage{times}
\usepackage{epsfig}
\usepackage{graphicx}
\usepackage{amsmath}
\usepackage{amssymb}
\usepackage{enumitem}
\usepackage{multirow}
\usepackage{booktabs}
\usepackage{pifont}
\usepackage{adjustbox}

\usepackage{algorithm}
\usepackage{algorithmic}

\usepackage{graphicx}
\usepackage{bm}
\usepackage{multirow}
\usepackage{caption}
\usepackage{xcolor}         % colors
\usepackage{siunitx}

\definecolor{cvprblue}{rgb}{0.21,0.49,0.74}
\usepackage[pagebackref,breaklinks,colorlinks,allcolors=cvprblue]{hyperref}

\def\paperID{750} % *** Enter the Paper ID here
\def\confName{CVPR}
\def\confYear{2026}

\title{Motion-Aware Animatable Gaussian Avatars Deblurring}

\author{
    Muyao Niu\textsuperscript{1,2,}\footnotemark[1] \quad
    Yifan Zhan\textsuperscript{1,2} \quad
    Qingtian Zhu\textsuperscript{1} \quad
    Zhuoxiao Li\textsuperscript{1} \quad 
    Wei Wang\textsuperscript{2} \quad \\[1pt]
    Zhihang Zhong\textsuperscript{3,}\textsuperscript{†}\quad
    Xiao Sun\textsuperscript{2} \quad
    Yinqiang Zheng\textsuperscript{1} \\[8pt]
    \hspace*{-2.5mm}\textsuperscript{1}The University of Tokyo
    \
    \textsuperscript{2}Shanghai Artificial Intelligence Laboratory
    \
    \textsuperscript{3}Shanghai Jiao Tong University \quad
}

\newcommand{\secondscore}[1]{\colorbox[rgb]{0.9686,0.8078,0.6274}{#1}}
\newcommand{\bestscore}[1]{\colorbox[rgb]{1,0.6,0.6}{#1}}
\newcommand{\thirdscore}[1]{\colorbox[rgb]{1,1,0.6509}{#1}}

\begin{document}

\maketitle

\renewcommand{\thefootnote}{\fnsymbol{footnote}}
\footnotetext[1]{Part of this work was done during the author's internship at the Shanghai Artificial Intelligence Laboratory.}
\renewcommand{\thefootnote}{\textsuperscript{†}}
\footnotetext[1]{Corresponding author.}

\renewcommand{\thefootnote}{1}

\begin{abstract}
The creation of 3D human avatars from multi-view videos is a significant yet challenging task in computer vision. However, existing techniques rely on high-quality, sharp images as input, which are often impractical to obtain in real-world scenarios due to variations in human motion speed and intensity. This paper introduces a novel method for directly reconstructing sharp 3D human Gaussian avatars from blurry videos. The proposed approach incorporates a 3D-aware, physics-based model of blur formation caused by human motion, together with a 3D human motion model designed to resolve ambiguities in motion-induced blur. This framework enables the joint optimization of the avatar representation and motion parameters from a coarse initialization. Comprehensive benchmarks are established using both a synthetic dataset and a real-world dataset captured with a 360-degree synchronous hybrid-exposure camera system. Extensive evaluations demonstrate the effectiveness of the model across diverse conditions. Codes Available: \url{https://github.com/MyNiuuu/MAD-Avatar}

\end{abstract}

\section{Introduction}
\label{sec:intro}

Motion blur arises when scene changes occur during camera exposure, degrading image quality and perceptual clarity. Despite advances in camera technology, it remains prevalent due to the unpredictable nature of object motion. Consequently, many methods have been proposed to restore sharp details from blurry captures~\cite{su2017deep,wang2019edvr,zamir2021multi,cao2022towards,zhong2023real,zhong2021towards}.

In the field of 3D reconstruction, creating high-quality 3D human avatars holds immense potential for the industry~\cite{jena2023splatarmor,wang2022arah,hu2024gauhuman,li2024animatable,cha2023generating,huang2024tech,liao2023high,zhang2024humanref,xu2025sequential,chen2024within,niu2025anicrafter,zhan2025r3}. Cutting-edge methods employing 3D Gaussian Splatting~\cite{kerbl20233d,gao2025proxy,liu2025maskgaussian,gao2025citygs} and the Skinned Multi-Person Linear (SMPL) model~\cite{loper2023smpl} have achieved notable success. These techniques~\cite{hu2024gauhuman,qian20243dgs,hu2024gaussianavatar,li2024animatable} typically rely on video data with SMPL parameters from calibration techniques like EasyMocap~\cite{peng2021neural}.

\begin{figure}[t]
    \centering
    \setlength{\abovecaptionskip}{2mm}
    \includegraphics[width=\linewidth]{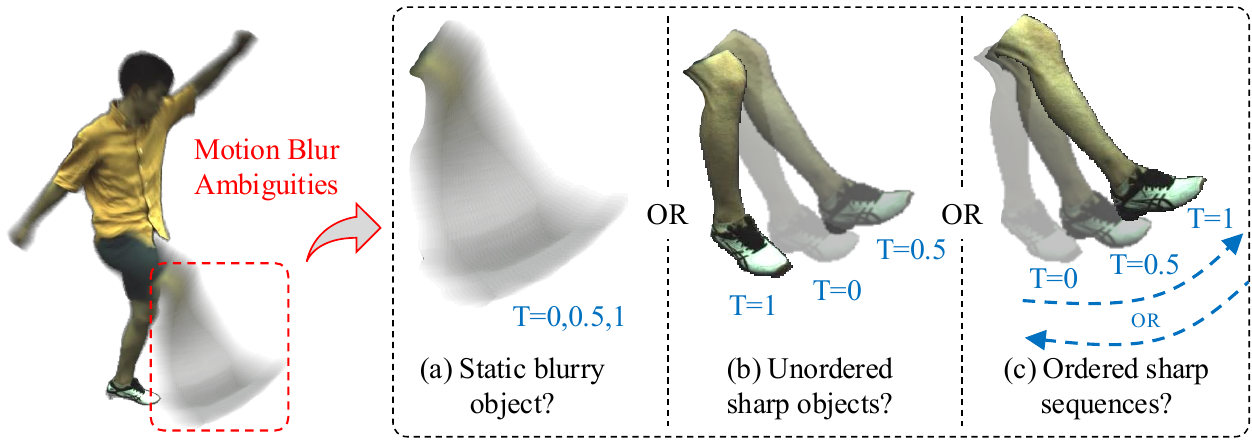}
    \caption{The ambiguity brought by motion blur. When reconstructing sharp 3DGS avatars from blurry frames, motion-induced blur introduces challenging ambiguities in motion interpretation.
    }
    \label{fig:disambiguate}
    \vspace{-2mm}
\end{figure}

While these models have achieved remarkable results with well-calibrated and high-quality video frames, their performance can be significantly degraded by blur effects caused by human motion. In practical scenarios, such motion-induced blur is often unavoidable due to the unpredictable speeds and movements of subjects.
In particular, blur effects can adversely affect the performance of existing models in two key ways. First, blurry captures may cause the 3DGS model to learn a distorted 3D representation, arising from the inherent ambiguity of motion blur (Fig.~\ref{fig:disambiguate}). Such ambiguity hinders the accurate recovery of structural information and texture details. Second, even though static cameras can be calibrated before recording, blurry captures still lead to erroneous estimation of SMPL parameters.\footnote{While certain approaches, such as GauHuman~\cite{hu2024gauhuman}, integrate pose refinement modules, they often yield suboptimal results because they lack a rigorous mechanism for modeling blur-aware motion.}

A practical strategy to mitigate this issue is to implement a two-stage baseline approach. First, 2D deblurring techniques~\cite{zamir2022restormer,zhong2023blur,pan2023deep,liang2024vrt} are applied to restore sharp video sequences. Then, the deblurred frames are used to train the 3DGS avatar model. While this baseline method improves visual quality by partially resolving motion blur ambiguity through 2D deblurring, it overlooks intrinsic 3D scene information. This omission often leads to inconsistencies across multiple views during the deblurring process, compromising the performance of the 3DGS avatar model.

Based on the settings of existing gaussian avatar models~\cite{hu2024gauhuman,qian20243dgs,hu2024gaussianavatar}, this paper introduces the first model for reconstructing sharp, animatable 3D human avatars directly from blurry video frames. By exploiting the distinctive characteristics of human avatar models, we expand the conventional physics-based 2D image blur process to a 3D-aware blur formation model. This formulation decomposes the inherently ill-posed deblurring problem into two primary tasks: optimizing sub-frame motion representations and constructing the canonical sharp 3DGS avatar model. To mitigate the sub-frame motion ambiguity introduced by motion-induced blur, a 3D-aware human motion model based on the SMPL framework is incorporated. This model enables the joint optimization of human motion and the sharp 3DGS avatar. Using the optimized motion and canonical 3DGS, motion-blurred frames are synthesized by averaging a sequence of ``virtual'' sharp images, and the loss is computed against the observed blurry frames.

Extensive evaluations are conducted on various datasets, baselines, and settings to confirm the superiority and robustness of the proposed method. As there are no blur-aware avatar benchmarks for qualitative and quantitative evaluation, a synthetic dataset is developed based on the widely used ZJU-MoCap dataset~\cite{peng2021neural}, complemented by a high-quality real-world dataset captured using a 360-degree hybrid-exposure system. A do-it-yourself style demo with iPhone 16 Pro is also provided. Codes and datasets will be publicly released to facilitate relevant research in the future.

\section{Related Work}
\label{sec:related}

\noindent \textbf{Video Deblurring.}
In recent years, deblurring has shifted from traditional methods~\cite{hyun2015generalized,levin2006blind,ren2017video,wulff2014modeling} to learning-based models that directly predict sharp videos~\cite{nah2017deep,tao2018scale,zhong2022animation,hyun2017online,nah2019recurrent,wang2022efficient,zhong2020efficient,zhou2019spatio,dai2017deformable,zhu2019deformable,pan2020cascaded,son2021recurrent,kupyn2018deblurgan,kupyn2019deblurgan,liang2021swinir,cao2022vdtr}. 
Recently, the advent of NeRF~\cite{mildenhall2021nerf} and 3DGS~\cite{kerbl20233d} has promoted multi-view deblurring. Specifically, methods like~\cite{ma2022deblur,wang2023bad,lee2023dp,zhao2024bad} focus on deblurring static scenes affected by defocus or camera movement, while some approaches~\cite{lu2025bard,sun2024dyblurf,bui2025moblurf,stearns2024dynamic} deblurs dynamic scenes. However, none of these methods could reconstruct sharp and animatable avatars from blurry videos.

\noindent \textbf{3D Human Avatars.}
Since the advent of Neural Radiance Fields~\cite{mildenhall2021nerf}, research on neural human rendering has flourished~\cite{noguchi2021neural,su2021nerf,xu2021h,guo2023vid2avatar,jiang2023instantavatar,jiang2022neuman,li2022tava,peng2021animatable,peng2022animatable,wang2022arah,weng2022humannerf,yu2023monohuman}.
Recently, the superior performance of 3DGS~\cite{kerbl20233d} has spurred extensive research into using 3D Gaussian representations for dynamic human reconstruction~\cite{zielonka2025drivable,li2024animatable,jena2023splatarmor,moreau2024human,ye2023animatable,kocabas2024hugs,lei2024gart,liu2024animatable,hu2024gaussianavatar}. 
However, these methods rely on high-quality input images and fail when applied to blurry captures.

\section{Method}
\label{sec:method}

\subsection{3D Blur Formation Model}

The physical mechanism of image formation involves capturing photons during the camera’s exposure period. This process can be mathematically represented in the 2D camera coordinate space as the integration of a sequence of conceptual sharp images over the exposure duration:
\begin{equation}
\mathbf{I}^B(\mathbf{u}) = \int^{\tau}_{0} \mathbf{I}^S_t(\mathbf{u}) \, \mathrm{d}t,
\end{equation}
where \(\mathbf{I}^B(\mathbf{u}) \in \mathbb{R}^{3 \times H \times W}\) denotes the captured blurry image, with \(H\) and \(W\) representing the image height and width, respectively. The variable \(\mathbf{u} \in \mathbb{R}^2\) indicates the pixel location, \(\tau\) is the exposure time, and \(\mathbf{I}^S_t(\mathbf{u}) \in \mathbb{R}^{3 \times H \times W}\) represents the virtual sharp image at time \(t\) within the exposure period. This continuous integration process can be discretely approximated as:
\begin{equation}
\mathbf{I}^B(\mathbf{u}) \approx \frac{1}{n} \sum^{n-1}_{i=0} \mathbf{I}^S_i(\mathbf{u}),
\end{equation}
where the blurred image \(\mathbf{I}^B(\mathbf{u})\) is estimated by averaging \(n\) virtual intermediate sharp images \(\mathbf{I}^S_i(\mathbf{u})\). The blur formation process is redefined from the perspective of 3D human avatar modeling, extending beyond the constraints of the 2D camera coordinate space. Specifically, the motion of a 3DGS avatar during the exposure period is modeled using a set of \(K\) 3D Gaussians \({G_k(\mathbf{x})}_{k=0}^{K-1}\) defined in the canonical space, together with a sequence of \(T\) SMPL~\cite{loper2023smpl} parameters \(\{\mathcal{S}_t\}^{T-1}_{t=0} = \{\bm{\Theta}_t, \bm{\beta}_t, \mathcal{B}_t\}^{T-1}_{t=0}\). These parameters dynamically deform the 3D Gaussians into the observation space at each discrete time step $t$. Consequently, the resulting blurry image \(\mathbf{I}^B\) can be expressed as:
\begin{equation}
\mathbf{I}^B = \frac{1}{T} \sum^{T-1}_{t=0} \mathcal{R}(\mathcal{W}(\{G_k(\mathbf{x})\}^{K-1}_{k=0}, \mathcal{S}_t), \bm{\mathrm{R}}, \bm{\mathrm{K}}),
\end{equation}
where \(\mathbf{x}\) denotes the coordinates of the 3D Gaussians, and \(\bm{\mathrm{R}}\) and \(\bm{\mathrm{K}}\) represent the camera’s extrinsic and intrinsic parameters, respectively. The operator \(\mathcal{R}\) denotes the rasterization process of the 3DGS model, while \(\mathcal{W}\) represents the warping of 3D Gaussians from the canonical space to the observation space, governed by the SMPL parameters \(\mathcal{S}_t\).

\subsection{3D Human Motion Model}
\label{sec:motion}

As illustrated in Fig.~\ref{fig:disambiguate}, motion-induced blur introduces significant ambiguities in motion that are readily observable in the pixel space. 
% Resolving this ill-posed problem requires accurately estimating plausible human motion throughout the exposure period. 
To tackle this challenge, a 3D-aware human motion model is proposed to effectively estimate both sub-frame motion within each exposure period and inter-frame global motion across consecutive frames.

\noindent \textbf{Sub-frame rigid sequential pose model.}
The pose parameters \(\bm{\Theta}_t\), defined in the SMPL model as \(\bm{\Theta}_t \in \mathbb{R}^{24 \times 3}\), represent the rotations of $24$ joints at each time step \(t\), expressed in \(\textbf{SO}(3)\). Considering the inherent continuity of joint motion, De Boor–Cox formulation for B-splines~\cite{qin1998general,farin2002curves,unser2002b,li2023usb,niu2024rs} is employed to interpolate intermediate poses.

For each joint \(j\) in the SMPL model, \(P\) control parameters are defined as \(\tilde{\bm{\Theta}}^j = \{\tilde{\bm{\Theta}}^j_p\}^{P-1}_{p=0} \in \mathbb{R}^{P \times 3}\), where \(P\) denotes the predefined number of control knots. These learnable parameters are initialized from the coarse estimation, and optimized during the training process. At each time step \(t\), the interpolation process begins with the computation of the timestep basis \(\mathbf{B}(t) \in \mathbb{R}^{1 \times P}\):
\begin{equation}
\mathbf{B}(t) = [1, \frac{t}{T}, (\frac{t}{T})^2, \ldots, (\frac{t}{T})^{P-1}],
\end{equation}
where \(T\) denotes the total exposure time. Based on this basis, the interpolated pose parameters for the \(j\)-th joint at time step \(t\) are computed as:
\begin{equation}
\hat{\bm{\Theta}}^j_t = \mathbf{B}(t) \cdot \mathcal{M}^{P} \cdot \tilde{\bm{\Theta}}^j,
\end{equation}
where \(\mathcal{M}^{P} \in \mathbb{R}^{P \times P}\) is the interpolation matrix~\cite{qin1998general}:
\begin{equation}
\mathcal{M}^{P}_{i,j} = C^{P-1}_{P-1-i} \sum_{s=j}^{P-1} (-1)^{s-j} C^{P}_{s-j} (P-s-1)^{P-1-i},
\end{equation}
where \(C^{n}_{k} = \frac{n!}{k!(n-k)!}\) denotes the binomial coefficient.

\begin{figure}
    \centering
    \includegraphics[width=\linewidth]{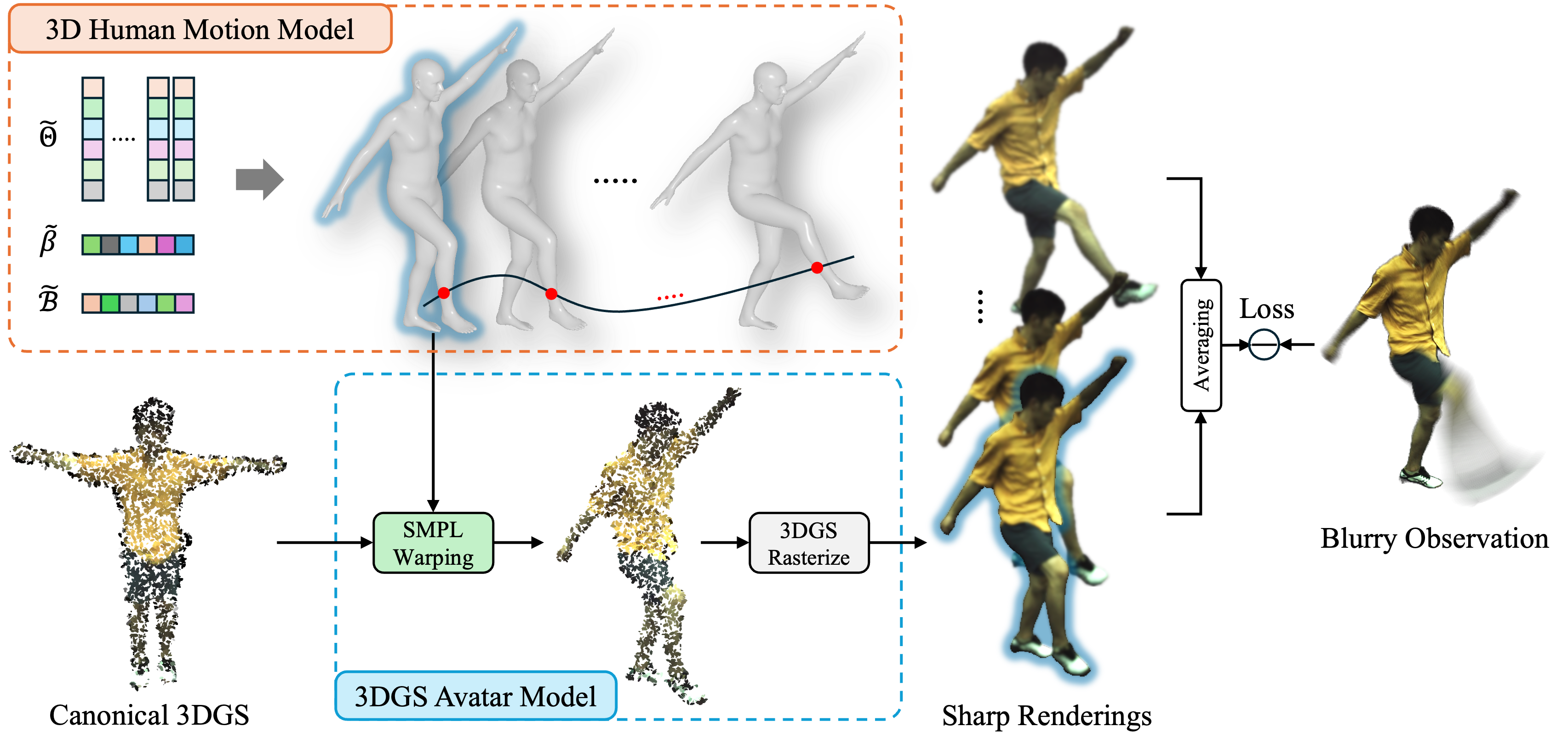}
    \caption{
    Brief illustration of the pipeline. 
    The sub-frame motion for each blurry frame is modeled using the SMPL representation, followed by warping the canonical 3DGS according to the estimated motion parameters. The final blurry image is synthesized by averaging the sequence of rendered virtual sharp images.
    }
    \label{fig:pipeline}
    % \vspace{-2mm}
\end{figure}

\begin{table*}[t]
\begin{minipage}[p]{0.64\textwidth}
\centering
\setlength{\abovecaptionskip}{2mm}
    \includegraphics[width=0.95\textwidth]{figures/data_capture_mosaic.jpg}
    \captionof{figure}{360-degree hybrid-exposure camera system. Left: the inner side and outer side of the capture cage. Right: illustration of the system and samples of captures. 
    }
    \label{fig:data_capture}
\end{minipage}
\hspace{0.01\textwidth}
\begin{minipage}[p]{0.30\textwidth}
\centering
\setlength{\abovecaptionskip}{2mm}
% \resizebox{\textwidth}{!}{%
\begin{tabular}{lcc}
% \hline
\hline
Capture type & Blur & Sharp \\
\hline
View number & $4$ & $8$ \\
Exposure & \SI{50}{ms} & \SI{3.125}{ms} \\
Resolution & \multicolumn{2}{c}{$2448 \times 2048$} \\
Camera model & \multicolumn{2}{c}{BFS-U3-51S5C} \\
Scene number & \multicolumn{2}{c}{$8$} \\
Picture format & \multicolumn{2}{c}{PNG} \\
% \hline
\hline
\end{tabular}%
% }
\caption{Configuration of the captured real data. 
}
\label{tab:data_capture}
\end{minipage}
\vspace{-1mm}
\end{table*}

\begin{figure*}[t]
\centering
\setlength{\abovecaptionskip}{2mm}
\includegraphics[width=\linewidth]{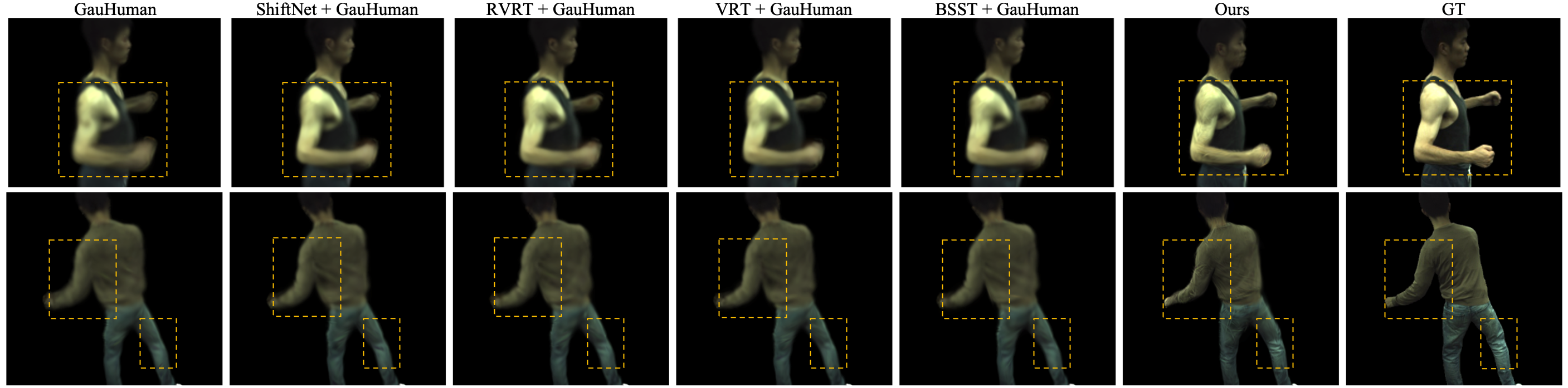}
\captionof{figure}{Qualitative comparison results on synthetic dataset. Zoom in for the best view.
}
\label{fig:synthetic}
\vspace{-2mm}
\end{figure*}

\noindent \textbf{Pose deformation model.}
While B-spline interpolation captures basic pose trajectories, it remains limited in representing non-rigid, high-frequency pose variations. To overcome this, a pose displacement \(\Delta^j_t\) is introduced for each joint \(j\) at every time step \(t\) within the exposure period:
\begin{align}
\Delta^j_t &= G_{disp}(\hat{\bm{\Theta}}^j_t; \theta_{disp}), \\
\bm{\Theta}^j_t &= \hat{\bm{\Theta}}^j_t + \Delta^j_t,
\end{align}
where \(G_{disp}\) denotes the CNN designed to estimate fine-grained pose variations, and \(\theta_{disp}\) represents its set of learnable parameters. This formulation allows the model to more accurately capture complex pose dynamics, restoring more realistic motion within each blurred frame.

\noindent \textbf{Inter-frame motion regularization.}
Although the L1 loss can yield satisfactory photometric results at the midpoint of the exposure period \((t=0.5)\), it may suffer from directional ambiguity, as illustrated in Fig.~\ref{fig:disambiguate}(c). This ambiguity occurs because motion in either direction can produce a similarly plausible blurry image, potentially leading to inaccurate motion direction estimation and misaligned renderings at non-midpoint timesteps. To mitigate this issue, a regularization term is introduced based on the inter-frame continuity commonly observed in video sequences. This term measures the Geodesic distance between the pose parameters at the final timestep of the current exposure period and those at the initial timestep of the subsequent exposure period:
\begin{equation}
    \mathcal{L}_{reg} = \frac{1}{24 \cdot (N_{e}-1)} \sum^{N_{e}-2}_{n=0}\sum^{23}_{j=0}\left|\hat{\mathbf{\Theta}}^j_{n,T-1} - \hat{\mathbf{\Theta}}^j_{n+1,0}\right|_{\text{G}},
\end{equation}
where \(\hat{\mathbf{\Theta}}^j_{n,t}\) denotes the estimated pose parameters of the \(j\)-th joint at timestep \(t\) for the \(n\)-th exposure period, and \(N_{e}\) represents the total number of exposure periods. $\left|\cdot\right|_{\text{G}}$ denotes the Geodesic distance. This regularization term is designed to enhance the inter-frame temporal coherence of joint movements, ensuring that the motion remains consistent and progresses naturally from one frame to the next.

\noindent \textbf{Shape estimation.} The shape parameters \(\hat{\bm{\beta}} \in \mathbb{R}^{10}\) of the SMPL model are initialized from the coarse estimation, then optimized as parameters during the training process.

\noindent \textbf{Linear Blend Skinning (LBS) weights estimation.} The SMPL model~\cite{loper2023smpl} provides pre-trained LBS weights, denoted as \(\tilde{\bm{\mathcal{B}}}\). Consistent with prior studies, identical LBS weights \(\hat{\bm{\mathcal{B}}}\) are employed across all time steps. Training begins with the initial pre-trained SMPL weights, which are further refined by introducing an LBS offset \(\delta\):
\begin{equation}
\hat{\bm{\mathcal{B}}} = \tilde{\bm{\mathcal{B}}} + \delta,
\end{equation}
where \(\delta\) is predicted by a simple CNN that takes the coordinates of all 3D Gaussians as input, enabling dynamic refinement of the LBS weights.

\subsection{Optimization}

\noindent \textbf{Pipeline.}
A brief illustration for the optimization pipeline of the proposed model is provided in Fig.~\ref{fig:pipeline}. The process begins with the estimation of sub-frame motion using the dedicated motion model described in Sec.~\ref{sec:motion}. The SMPL parameters are initialized using the coarse estimation from blurry frames, and gradually optimized within the 3D-aware framework. The canonical 3D Gaussian Splatting (3DGS) representation is then transformed according to the estimated motion parameters. Subsequently, sharp virtual images are rendered for each intermediate timestep. These images are averaged to produce the final blurry image used for loss computation.

\begin{figure}[t]
\vspace{-1mm}
\centering
\setlength{\abovecaptionskip}{2mm}
\includegraphics[width=\linewidth]{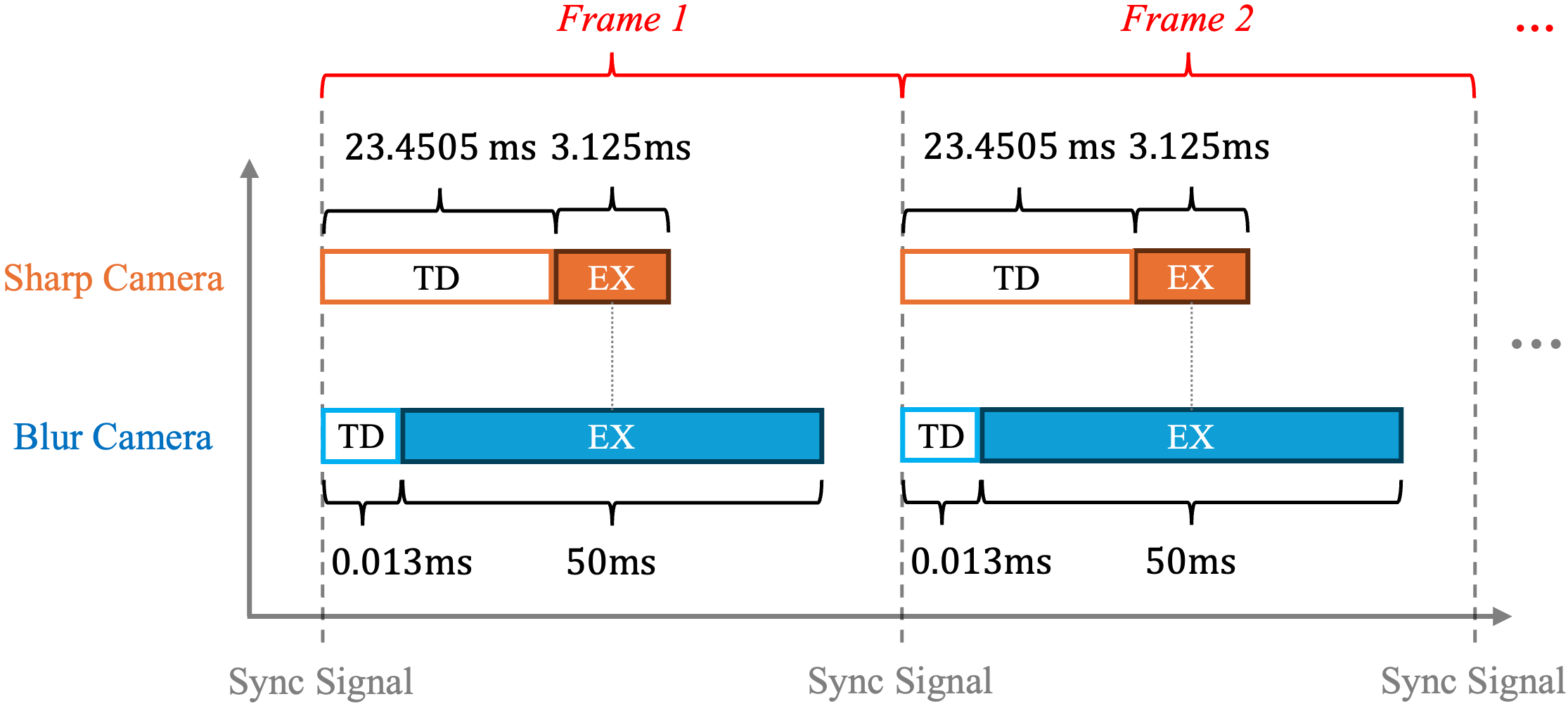}
\caption{Time synchronization of the camera system. `TD' and `EX' stand for ``Trigger Delay'' and ``Exposure''.}
\label{fig:time_sync}
\vspace{-2mm}
\end{figure}

\noindent \textbf{Loss function.}
The loss function includes the L1 loss between the synthesized blurry frame \(\hat{\mathbf{I}}^B\) and the observed blurry frame \(\mathbf{I}^B\), and the inter-frame motion regularization loss as decribed in Sec.~\ref{sec:motion}:
\begin{equation}
    \mathcal{L} = ||\hat{\mathbf{I}}^B - \mathbf{I}^B||_1 + \mathcal{L}_{reg}.
\end{equation}
The total loss enforces both accurate avatar reconstruction and the estimation of realistic, temporally coherent motion.

\section{Experiments}
\label{sec:experiments}

\begin{figure*}[t]
\centering
\setlength{\abovecaptionskip}{2mm}
\includegraphics[width=\linewidth]{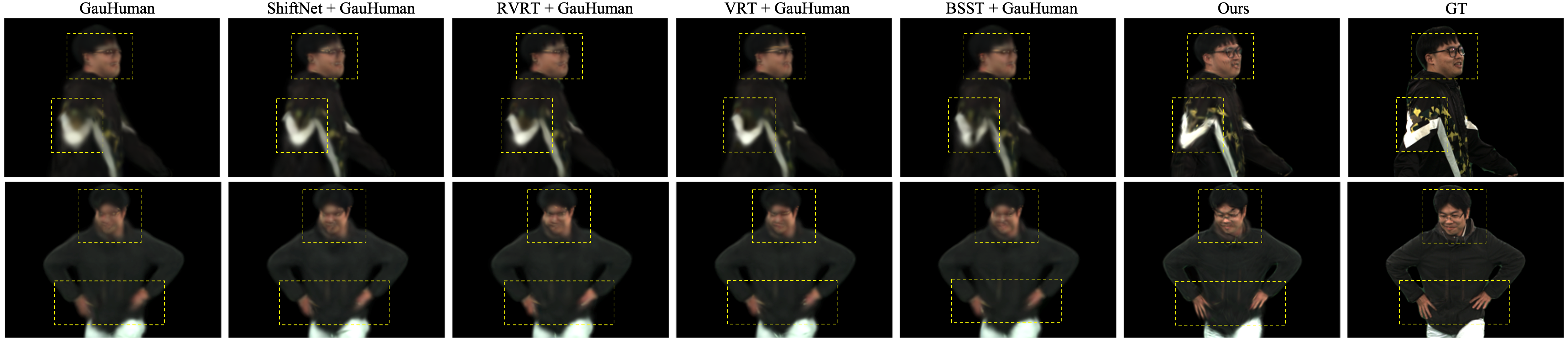}
\captionof{figure}{Qualitative comparison results on real data. Zoom in for the best view.
}
\label{fig:real_data}
\vspace{-1mm}
\end{figure*}

\begin{figure*}
    \centering
    \setlength{\abovecaptionskip}{2mm}
    \includegraphics[width=\linewidth]{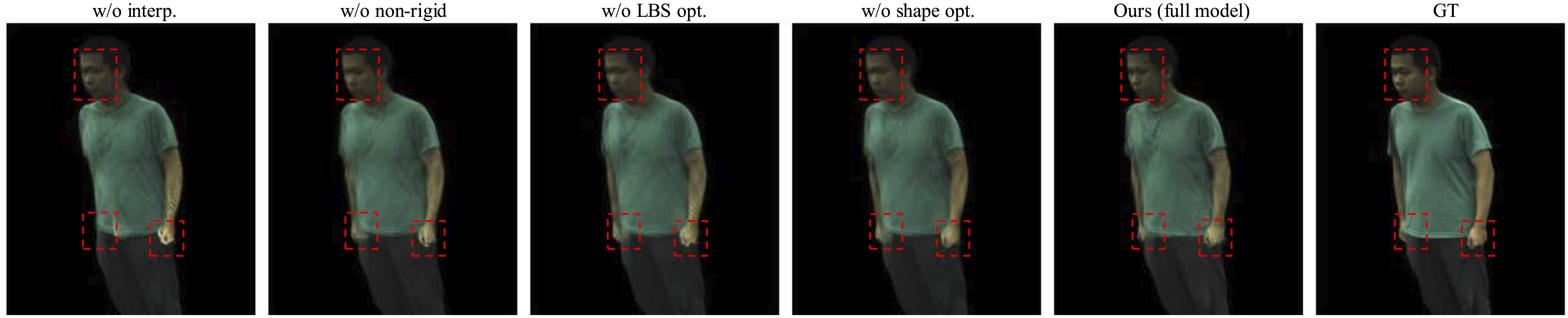}
    \caption{Qualitative results for ablation studies. Zoom in for the best view.}
    \label{fig:ablation}
    % \vspace{-1mm}
\end{figure*}

\subsection{Benchmarks}

\noindent \textbf{Synthetic dataset.}
The synthetic data is constructed based on the ZJU-MoCap~\cite{peng2021neural} dataset since it is widely adopted by many 3D human avatar research work~\cite{hu2024gauhuman,geng2023learning,weng2022humannerf,qian20243dgs,kim2024motion,yin2025event}. Following their practice, six human subjects (IDs: $377$, $386$, $387$, $392$, $393$, $394$) are selected for experiments. 
Each blurry frame was synthesized by averaging $K_{blur}$ consecutive sharp frames
To mitigate discontinuities caused by direct averaging, the RIFE model~\cite{huang2022real} is employed to interpolate $16$ intermediate frames between adjacent sharp frames. Blurry frames from Cameras $04$, $10$, $16$, and $22$ are used for training, while the sharp frames from the remaining $18$ cameras are reserved for novel-view test. SMPL parameters are calibrated using the blurry video with EasyMoCap~\cite{peng2021neural}, and human masks are obtained using SAM~\cite{kirillov2023segment}.

\begin{table}[t]
\centering
\setlength{\tabcolsep}{2pt}
\setlength{\abovecaptionskip}{2mm}
\caption{Quantitative comparison results on two datasets. 
We colorize results as \bestscore{best}, \secondscore{second best}, and \thirdscore{third best}. 
}
\label{tab:synthetic_novel}
\resizebox{\linewidth}{!}{%
\begin{tabular}{lcccccc}
\hline
\multirow{2}{*}{Methods} & \multicolumn{3}{c}{Synthetic Dataset} & \multicolumn{3}{c}{Real Dataset} \\
& PSNR $\uparrow$ & SSIM $\uparrow$ & LPIPS $\downarrow$ & PSNR $\uparrow$ & SSIM $\uparrow$ & LPIPS $\downarrow$ \\ \hline
GauHuman & 23.080 & 0.7660 & 0.2277 & \secondscore{25.602} & 0.8044 & 0.2380 \\
ShiftNet + GauHuman & \secondscore{23.089} & 0.7695 & 0.2219 & 25.549 & 0.8043 & 0.2347 \\
RVRT + GauHuman & 23.078	& \thirdscore{0.7697} & 0.2218 & 25.547 & 0.8065 & \thirdscore{0.2343} \\
VRT + GauHuman & 23.074 & 0.7696 & \secondscore{0.2205} & 25.553 & \thirdscore{0.8067} & 0.2345 \\
BSST + GauHuman & \thirdscore{23.081} & \secondscore{0.7698}	& \thirdscore{0.2212} & \thirdscore{25.568} & \secondscore{0.8068} & \secondscore{0.2342} \\
Ours & \bestscore{25.546} & \bestscore{0.8290} & \bestscore{0.1476} & \bestscore{27.010} & \bestscore{0.8271} & \bestscore{0.1668} \\ \hline
\end{tabular}%
}
% \vspace{-2mm}
\end{table}

\begin{table}[t]
\centering
\setlength{\tabcolsep}{4pt}
\setlength{\abovecaptionskip}{2mm}
\caption{Quantitative ablation results on two datasets. We colorize results as \bestscore{best}, \secondscore{second best}, and \thirdscore{third best}.}
\label{tab:ablation}
\resizebox{\linewidth}{!}{%
\begin{tabular}{lcccccc}
\hline
\multirow{2}{*}{Models} & \multicolumn{3}{c}{Synthetic Dataset} & \multicolumn{3}{c}{Real Dataset} \\
& PSNR $\uparrow$ & SSIM $\uparrow$ & LPIPS $\downarrow$ & PSNR $\uparrow$ & SSIM $\uparrow$ & LPIPS $\downarrow$ \\ \hline
w/o interp. & 24.009 & 0.8053 & 0.1620 & 25.825 & 0.8140 & 0.1729 \\
w/o pose deform & 25.301 & 0.8229 & 0.1545 & 26.426 & 0.8184 & 0.1743 \\
w/o LBS opt. & \thirdscore{25.394} & \thirdscore{0.8261} & \thirdscore{0.1486} & \thirdscore{26.821} & \thirdscore{0.8233} & \thirdscore{0.1697} \\
w/o shape opt. & \secondscore{25.529} & \secondscore{0.8284} & \secondscore{0.1481} & \secondscore{26.964} & \secondscore{0.8261} & \secondscore{0.1669} \\
Ours (full model) & \bestscore{25.546} & \bestscore{0.8290} & \bestscore{0.1476} & \bestscore{27.010} & \bestscore{0.8271} & \bestscore{0.1668} \\ 
\hline
\end{tabular}%
}
% \vspace{-2mm}
\end{table}

\begin{figure}
    \vspace{-1mm}
    \centering
    \setlength{\abovecaptionskip}{2mm}
    \includegraphics[width=\linewidth]{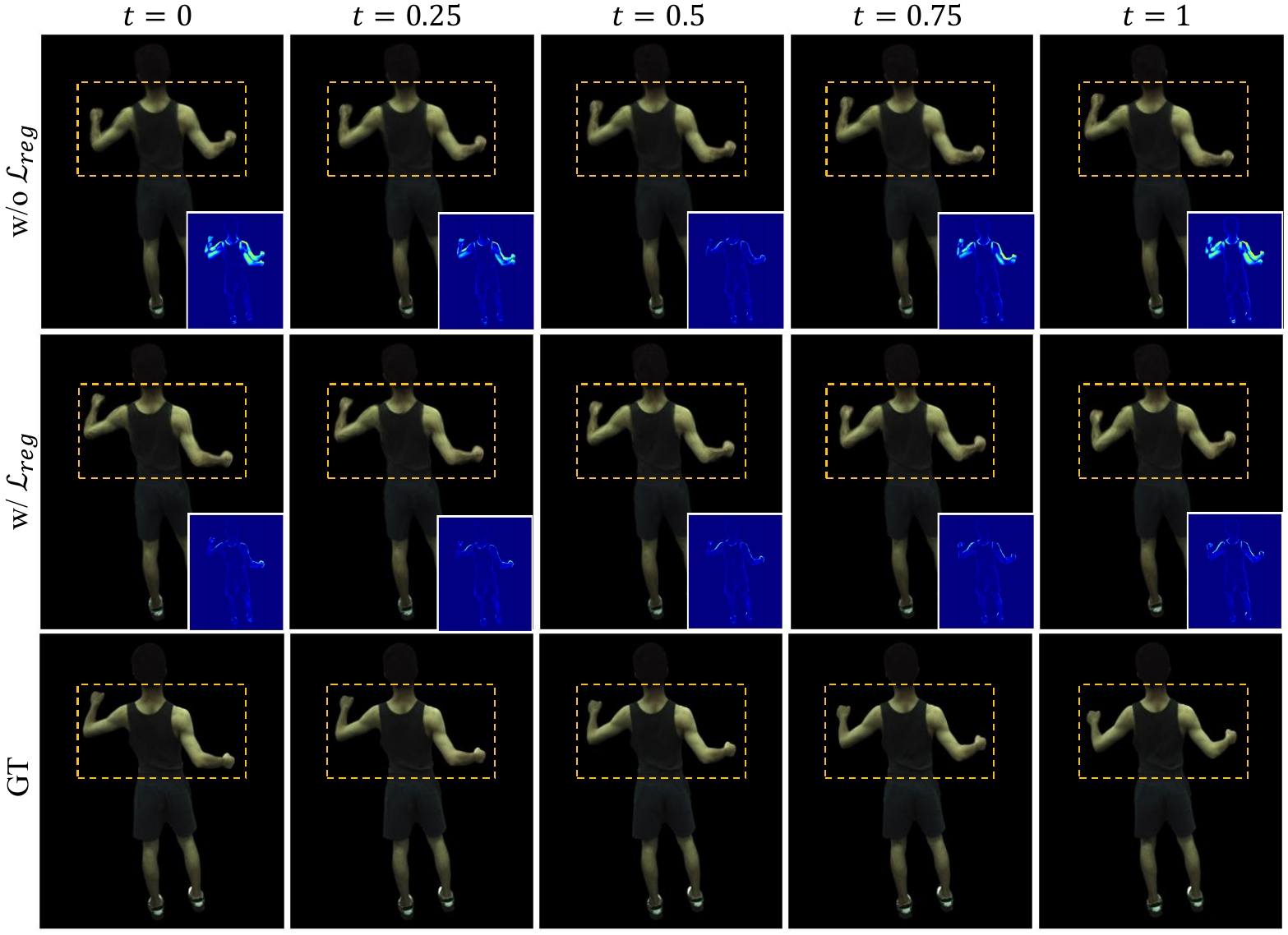}
    \caption{Qualitative ablation results for $\mathcal{L}_{reg}$. Model w/o $\mathcal{L}_{reg}$ gives worse result for non-middle timesteps due to incorrect motion direction estimates. Error maps are visualized for each output.}
    \label{fig:ablation_Lreg}
    % \vspace{-1mm}
\end{figure}

\noindent \textbf{Real dataset.}
The dataset consists of real-captured blurry videos and synchronously captured sharp videos from novel viewpoints. The capture system employs $12$ time-synchronized Blackfly BFS-U3-51S5C cameras arranged in a 360-degree spatial configuration, as illustrated in Fig.~\ref{fig:data_capture}. $4$ cameras record blurry videos, while the remaining $8$ capture sharp videos for novel-view evaluation. Fig.~\ref{fig:time_sync} illustrates the time synchronization of two types of cameras. 
To ensure consistent pixel intensity, the blurry cameras are equipped with a ${1}/{16}$ ND filter. 
Detailed configuration are reported in Tab.~\ref{tab:data_capture}.
SMPL parameters are calibrated using EasyMoCap, and human masks are generated using SAM~\cite{kirillov2023segment}.

\begin{table}[t]
\centering
\setlength{\tabcolsep}{3pt}
\setlength{\abovecaptionskip}{2mm}
\caption{Quantitative ablation results for the inter-frame motion regularization loss $\mathcal{L}_{reg}$. We colorize \thirdscore{better} result of each sector.}
\label{tab:Lreg}
\resizebox{\linewidth}{!}{%
\begin{tabular}{llcccccc}
\hline
\multirow{2}{*}{$K_{blur}$} & \multirow{2}{*}{Models} & \multicolumn{3}{c}{Middle Timestep} & \multicolumn{3}{c}{Non-middle Timesteps} \\
 &  & PSNR $\uparrow$ & SSIM $\uparrow$ & LPIPS $\downarrow$ & PSNR $\uparrow$ & SSIM $\uparrow$ & LPIPS $\downarrow$ \\
 \hline
\multirow{2}{*}{$5$} & w/o $\mathcal{L}_{reg}$ & \thirdscore{25.567} & \thirdscore{0.8296} & {0.1478} & {24.421} & {0.8111} & {0.1601} \\
 & w/ $\mathcal{L}_{reg}$ & {25.546} & {0.8290} & \thirdscore{0.1476} & \thirdscore{25.417} & \thirdscore{0.8269} & \thirdscore{0.1485} \\
  \hline
\multirow{2}{*}{$7$} & w/o $\mathcal{L}_{reg}$ & {25.113} & {0.8197} & {0.1585} & {23.737} & {0.7950} & {0.1758}  \\
 & w/ $\mathcal{L}_{reg}$ & \thirdscore{25.155} & \thirdscore{0.8200} & \thirdscore{0.1557} & \thirdscore{25.036} & \thirdscore{0.8179} & \thirdscore{0.1567} \\
  \hline
\multirow{2}{*}{$9$} & w/o $\mathcal{L}_{reg}$ & {24.628} & {0.8114} & {0.1680} & {23.198} & {0.7825} & {0.1888}  \\
 & w/ $\mathcal{L}_{reg}$ & \thirdscore{24.680} & \thirdscore{0.8126} & \thirdscore{0.1636} & \thirdscore{24.605} & \thirdscore{0.8111} & \thirdscore{0.1645} \\
\hline
\multirow{2}{*}{$11$} & w/o $\mathcal{L}_{reg}$ & {24.241} & {0.8009} & {0.1786} & {22.926} & {0.7717} & {0.1998} \\
 & w/ $\mathcal{L}_{reg}$ & \thirdscore{24.353} & \thirdscore{0.8039} & \thirdscore{0.1725} & \thirdscore{24.317} & \thirdscore{0.8031} & \thirdscore{0.1726} \\
\hline
\end{tabular}%
}
% \vspace{-1mm}
\end{table}

\noindent \textbf{Implementation details.}
Adam optimizer~\cite{kingma2014adam} is used with parameters \(\beta_1 = 0.9\) and \(\beta_2 = 0.999\). The learning rates and decay schedules follow the original 3DGS~\cite{kerbl20233d}. Following the protocol in~\cite{hu2024gauhuman}, the input resolution is set to \(512 \times 512\) for the synthetic dataset and \(612 \times 512\) for the real dataset. 
All models are trained on a single NVIDIA GeForce RTX 4090 GPU. 
Since the images contain large black regions, each image is cropped to the human bounding box for quantitative evaluation.

\subsection{Comparisons}

The proposed model is compared against several baseline approaches to assess its effectiveness. The simplest baseline involves training a SOTA avatar model (\eg, GauHuman~\cite{hu2024gauhuman}) using blurry video frames. A two-stage baseline is then considered, where 2D deblurring methods (RVRT~\cite{liang2022recurrent}, ShiftNet~\cite{li2023simple}, VRT~\cite{liang2024vrt}, and BSST~\cite{zhang2024blur}) are first applied to restore sharp frames, then used to train the avatar model.

\noindent \textbf{Quantitative results.}
Quantitative comparisons on two datasets are presented in Tab.~\ref{tab:synthetic_novel}. The two-stage baselines exhibit suboptimal performance, with noticeably lower quantitative metrics. This limitation arises because 2D deblurring models fail to maintain multi-view consistency across different viewpoints and lack explicit modeling of the scene’s intrinsic 3D structure. They only slightly outperform the direct 3DGS avatar model, highlighting that inconsist 2D deblurring constrains the 3DGS model’s ability to produce accurate reconstructions. In contrast, the proposed method outperforms baselines by explicitly modeling the 3D motion-induced blur formation process.

\begin{figure}[t]
\centering
\setlength{\abovecaptionskip}{2mm}
\includegraphics[width=.95\linewidth]{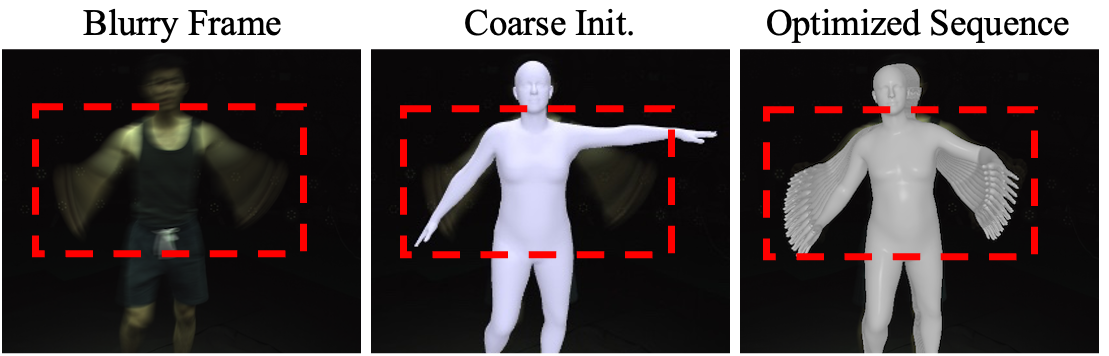}
\captionof{figure}{Visualization of the initial estimated SMPL by EasyMocap and optimized SMPL sequence by the proposed model.}
\label{fig:viz_mesh}
% \vspace{-2mm}
\end{figure}

\begin{table}[t]
\centering
\setlength{\tabcolsep}{5pt}
\setlength{\abovecaptionskip}{2mm}
\caption{Qualitative ablation results on trajectories representations. We colorize result as \bestscore{best}, \secondscore{second best}, and \thirdscore{third best}.}
\label{tab:traj_rep}
\resizebox{\linewidth}{!}{%
\begin{tabular}{lcccccc}
\hline
\multirow{2}{*}{Models} & \multicolumn{3}{c}{Synthetic Dataset} & \multicolumn{3}{c}{Real Dataset} \\
& PSNR $\uparrow$ & SSIM $\uparrow$ & LPIPS $\downarrow$ & PSNR $\uparrow$ & SSIM $\uparrow$ & LPIPS $\downarrow$ \\ 
\hline
Linear & \thirdscore{25.516} & \thirdscore{0.8289} & \thirdscore{0.1483} & \thirdscore{26.899} & \thirdscore{0.8257} & \thirdscore{0.1680} \\
Slerp & \secondscore{25.539} & \secondscore{0.8288} & \secondscore{0.1481} & \bestscore{27.013} & \secondscore{0.8264} & \secondscore{0.1674} \\
B-Spline & \bestscore{25.546} & \bestscore{0.8290} & \bestscore{0.1476} & \secondscore{27.010} & \bestscore{0.8271} & \bestscore{0.1668} \\ 
\hline
\end{tabular}%
    }
% \vspace{-2mm}
\end{table}

\begin{table}[t]
\centering
\setlength{\tabcolsep}{6pt}
\setlength{\abovecaptionskip}{2mm}
\caption{Ablation results for control knot number $P$.}
\label{tab:hyper_p}
\resizebox{.9\linewidth}{!}{%
\begin{tabular}{lcccccc}
\hline
\multirow{2}{*}{$P$} & \multicolumn{3}{c}{Synthetic Dataset} & \multicolumn{3}{c}{Real Dataset} \\
 & PSNR $\uparrow$ & SSIM $\uparrow$ & LPIPS $\downarrow$ & PSNR $\uparrow$ & SSIM $\uparrow$ & LPIPS $\downarrow$ \\ \hline
$2$ & \thirdscore{25.516} & \secondscore{0.8289} & \secondscore{0.1480} & 26.899 & 0.8257 & 0.1680 \\
$3$ & 25.482 & 0.8284 & \thirdscore{0.1485} & \secondscore{27.003} & \secondscore{0.8266} & \secondscore{0.1671} \\
$4$ & \bestscore{25.546} & \bestscore{0.8290} & \bestscore{0.1476} & \bestscore{27.010} & \bestscore{0.8271} & \bestscore{0.1668} \\
$5$ & \secondscore{25.540} & \thirdscore{0.8288} & 0.1490 & \thirdscore{26.977} & \thirdscore{0.8260} & \secondscore{0.1673} \\ \hline
\end{tabular}%
}
% \vspace{-2mm}
\end{table}

\noindent \textbf{Qualitative results.}
Qualitative comparison results are shown in Fig.~\ref{fig:synthetic} and Fig.~\ref{fig:real_data}. While integrating 2D deblurring techniques with the 3DGS avatar model provides minor improvements in reconstruction quality (\eg, the arm region in Fig.~\ref{fig:synthetic}), prominent visual artifacts remain, including residual blur and loss of detail around body contours. These artifacts arise from inconsistencies in deblurring across multiple viewpoints, as 2D deblurring methods do not enforce multi-view consistency. The proposed approach incorporates a physics-based blur formation model, allowing accurate learning of the 3D representation and enabling simultaneous deblurring and improved avatar reconstruction.

\subsection{Ablation Study}

\noindent \textbf{Model components.}
Ablation studies are conducted to evaluate the contributions of different model components.
\begin{itemize}
\item \textbf{w/o interp.}: Independently optimizing the pose at each timestep instead of using pose interpolation. 
\item \textbf{w/o pose deform}: Using only the rigid sequential pose model, excluding pose deformation model. 
\item \textbf{w/o LBS opt.}: Employing pre-trained LBS weights without additional optimization. 
\item \textbf{w/o shape opt.}: Directly using calibrated SMPL shape parameters without further refinement.
\end{itemize}
The quantitative and qualitative results are presented in Tab.~\ref{tab:ablation} and Fig.~\ref{fig:ablation}. Independently optimizing poses for each intermediate timestep leads to unordered motion estimations (the case illustrated in Fig.~\ref{fig:disambiguate}~(b)), leading to misaligned deblurring results. Excluding the pose deformation model introduces additional artifacts, as B-spline interpolation alone cannot adequately capture complex human motion. The performance also degrades when LBS weights and SMPL shape parameters are not refined, emphasizing the necessity of these components.

\begin{figure}[t]
\vspace{-1mm}
\centering
\setlength{\abovecaptionskip}{2mm}
\includegraphics[width=.8\linewidth]{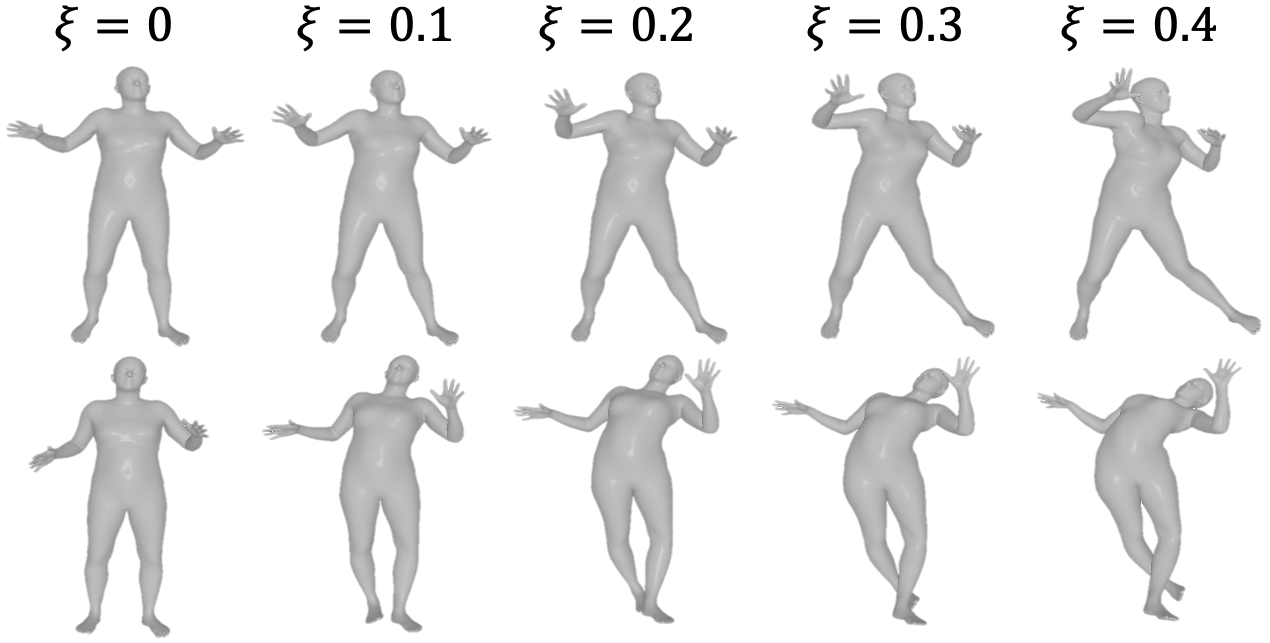}
\captionof{figure}{Visualizing SMPL under different perturbations.}
\label{fig:viz_perturb}
% \vspace{-2mm}
\end{figure}

\begin{table}[t]
\centering
\setlength{\tabcolsep}{6pt}
\setlength{\abovecaptionskip}{2mm}
\caption{Ablation results for virtual sharp image number $T$.}
\label{tab:hyper_t}
\resizebox{.9\linewidth}{!}{%
\begin{tabular}{lcccccc}
\hline
\multirow{2}{*}{$T$} & \multicolumn{3}{c}{Synthetic Dataset} & \multicolumn{3}{c}{Real Dataset} \\
 & PSNR $\uparrow$ & SSIM $\uparrow$ & LPIPS $\downarrow$ & PSNR $\uparrow$ & SSIM $\uparrow$ & LPIPS $\downarrow$ \\ \hline
$3$ & \thirdscore{25.528} & 0.8276 & 0.1503 & 26.837 & 0.8253 & 0.1695 \\
$5$ & \bestscore{25.546} & \bestscore{0.8290} & \bestscore{0.1476} & \bestscore{27.010} & \bestscore{0.8271} & \bestscore{0.1668} \\
$7$ & \secondscore{25.532} & \secondscore{0.8281} & \secondscore{0.1499} & \secondscore{26.993} & \secondscore{0.8269} & \secondscore{0.1670} \\
$9$ & 25.522 & \thirdscore{0.8279} & \thirdscore{0.1501} & \thirdscore{26.965} & \thirdscore{0.8261} & \thirdscore{0.1676} \\ \hline
\end{tabular}%
}
% \vspace{-2mm}
\end{table}

\begin{table}[t]
\centering
\setlength{\abovecaptionskip}{2mm}
\caption{Quantitative results with different perturbations.}
\label{tab:perterbation}
\resizebox{0.7\linewidth}{!}{%
\begin{tabular}{lccccc}
\hline
$\xi$ & $0$ & $0.1$ & $0.2$ & $0.3$ & $0.4$ \\
\hline
PSNR & \bestscore{25.55} & \secondscore{25.51} & \thirdscore{25.44} & 25.39 & 25.14 \\
SSIM & \bestscore{.8290} & \secondscore{.8284} & \thirdscore{.8271} & .8258 & .8225 \\
LPIPS & \bestscore{.1476} & \bestscore{.1476} & \thirdscore{.1493} & .1504 & .1524 \\
\hline
\end{tabular}%
}
% \vspace{-2mm}
\end{table}

\noindent \textbf{Inter-frame regularization loss.}
Evaluations on $\mathcal{L}_{reg}$ are summarized in Tab.~\ref{tab:Lreg} and Fig.~\ref{fig:ablation_Lreg}. It improves performance especially for non-middle timesteps by enabling more accurate estimations for inter-frame motion directions. Quantitative results on real data are excluded due to the lack of GT frames for non-middle timesteps.

\noindent \textbf{Trajectory representations.}
Different trajectory representations are evaluated in the proposed model, including B-spline interpolation, linear interpolation, and spherical linear interpolation (Slerp). The corresponding results are presented in Tab.~\ref{tab:traj_rep}. It can be observed that B-spline interpolation generally achieves the best performance.

\noindent \textbf{Hyperparameters.}
The proposed model is evaluated with varying numbers of control knots, \(P = 2, 3, 4, 5\), for B-spline sequential pose interpolation. As reported in Tab.~\ref{tab:hyper_p}, the best performance is obtained with \(P = 4\). However, the performance differences remain relatively small, as the  pose deformation model enhances robustness for smaller \(P\) values by effectively modeling complex motions, thereby reducing performance gaps across different settings of \(P\).
% \noindent \textbf{Impact of virtual sharp image number $T$.}
The influence of virtual sharp image number \(T\) is examined in Tab.~\ref{tab:hyper_t}. Optimal performance is achieved when \(T = 5\).

\subsection{Robustness and Generality Evaluation}
\label{sec:robust_smpl}

\begin{figure}[t]
    \centering
    \setlength{\abovecaptionskip}{2mm}
    \includegraphics[width=\linewidth]{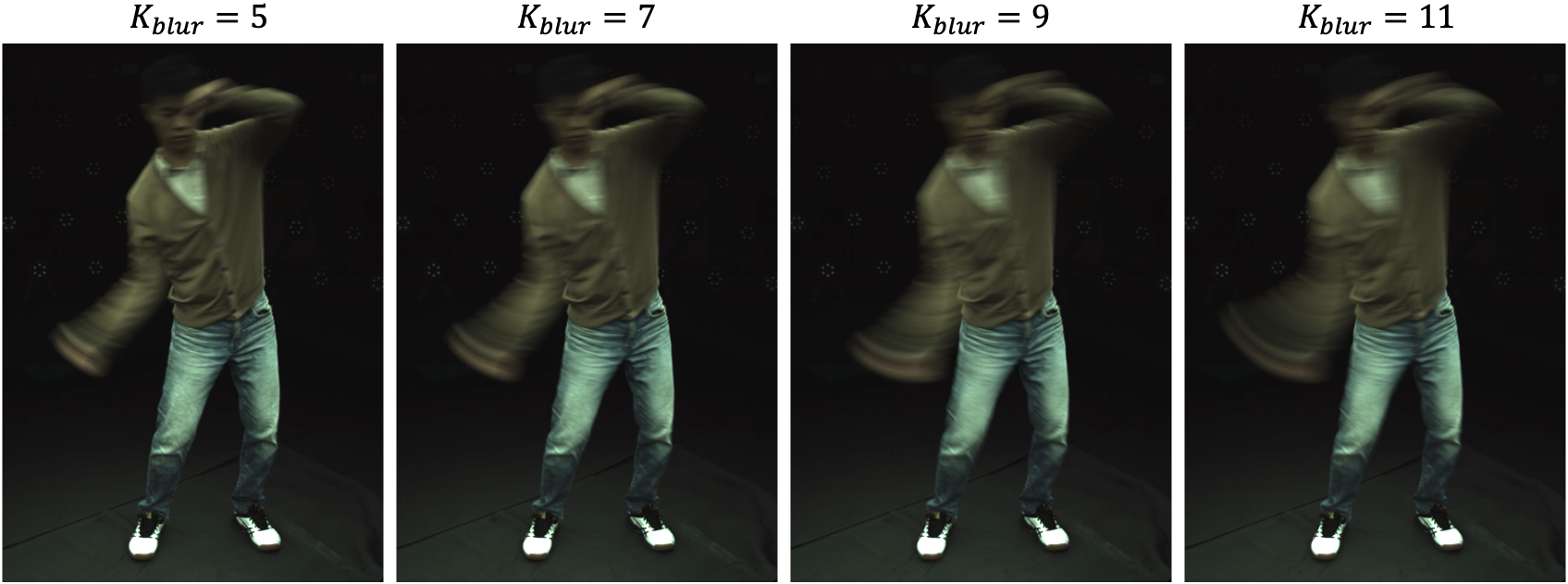}
    \caption{Visualization for different blur magnitudes $K_{blur}$.}
    \label{fig:K_blur}
    % \vspace{-1mm}
\end{figure}

\begin{table}[t]
\centering
\setlength{\abovecaptionskip}{2mm}
\caption{Quantitative result w/o optimizing SMPL parameters.}
\setlength{\tabcolsep}{1mm}
\resizebox{\linewidth}{!}{
\begin{tabular}{lcccccc}
\hline
\multirow{2}{*}{Methods} & \multicolumn{3}{c}{Synthetic Dataset} & \multicolumn{3}{c}{Real Dataset} \\
& PSNR $\uparrow$ & SSIM $\uparrow$ & LPIPS $\downarrow$ & PSNR $\uparrow$ & SSIM $\uparrow$ & LPIPS $\downarrow$ \\ \hline
w/o SMPL opt. & 21.63 & .7269 & .2421 & 25.14 & .7855 & .2423 \\
Ours & {25.55} & {.8290} & {.1476} & {27.01} & {.8271} & {.1668}\\
\hline
\end{tabular}%
}
\label{tab:no_pose_opt}
% \vspace{-1mm}
\end{table}

\begin{table}[t]
\centering
\setlength{\abovecaptionskip}{2mm}
\caption{Quantitative comparisons with different $K_{blur}$.
}
\label{tab:K_blur}
\resizebox{\linewidth}{!}{%
\begin{tabular}{llcccccc}
\hline
\multirow{2}{*}{$K_{blur}$} & \multirow{2}{*}{Metrics} & \multirow{2}{*}{GauH} & ShiftNet & RVRT & VRT & BSST & \multirow{2}{*}{Ours} \\
 & & & + GauH & + GauH & + GauH & + GauH & \\
\hline
\multirow{3}{*}{$7$} & PSNR $\uparrow$ & 22.983 & \thirdscore{23.003} & 22.992 & 22.972 & \secondscore{23.009} & \bestscore{25.155} \\
 & SSIM $\uparrow$ & 0.7599 & 0.7648 & 0.7650 & \secondscore{0.7654} & \thirdscore{0.7652} & \bestscore{0.8200} \\
 & LPIPS $\downarrow$ & 0.2378 & 0.2314 & 0.2301 & \thirdscore{0.2299} & \secondscore{0.2294} & \bestscore{0.1557} \\
\hline
\multirow{3}{*}{$9$} & PSNR $\uparrow$ & 22.693 & \thirdscore{22.699} & 22.692 & 22.693 & \secondscore{22.732} & \bestscore{24.680} \\
 & SSIM $\uparrow$ & 0.7532 & 0.7585 & 0.7587 & \thirdscore{0.7588} & \secondscore{0.7590} & \bestscore{0.8126} \\
 & LPIPS $\downarrow$ & 0.2497 & 0.2426 & \thirdscore{0.2411} & 0.2412 & \secondscore{0.2409} & \bestscore{0.1636} \\
\hline
\multirow{3}{*}{$11$} & PSNR $\uparrow$ & 22.635 & 22.632 & 22.644 & \thirdscore{22.647} & \secondscore{22.652} & \bestscore{24.353} \\
 & SSIM $\uparrow$ & 0.7549 & 0.7554 & 0.7555 & \secondscore{0.7564} & \thirdscore{0.7557} & \bestscore{0.8039} \\
 & LPIPS $\downarrow$ & 0.2552 & 0.2502 & \secondscore{0.2477} & \thirdscore{0.2480} & 0.2482 & \bestscore{0.1725} \\
\hline
\end{tabular}%
}
% \vspace{-1mm}
\end{table}

\noindent \textbf{Influence of the accuracy of SMPL estimation.} SMPL estimations from blurry data could be inaccurate. However, the proposed approach does not depend on precise SMPL estimates but only coarse initializations. Fig.~\ref{fig:viz_mesh} visualizes the initial and optimized SMPL parameters, showing that the method refines inaccurate initializations and recover accurate sub-frame SMPL poses. 
To further assess robustness against inaccurate initialization, random perturbations $\epsilon \sim U[-\xi, \xi]$ are added to the initializations. Fig.~\ref{fig:viz_perturb} visualizes the effects of varying $\xi$ values. The quantitative results in Tab.~\ref{tab:perterbation} confirm the method’s resilience to even large perturbations. Finally, experiments are conducted without optimizing SMPL (\textit{pose, shape, LBS}), using only the initial estimates. The results in Tab.~\ref{tab:no_pose_opt} proves the necessity of SMPL optimization.

\noindent \textbf{Different blur magnitudes $K_{blur}$.}
To evaluate the robustness of the proposed method under varying blur intensities, additional datasets were synthesized with blur magnitudes \(K_{blur} = 7, 9, 11\). Visualization of different blur magnitudes $K_{blur}$ is provided in Fig.~\ref{fig:K_blur}. As reported in Tab.~\ref{tab:K_blur}, the proposed method consistently outperforms all baseline approaches across varying blur level. Furthermore, corresponding ablation results in Tab.~\ref{tab:ablation_Kblur} confirm the contribution of each model component under different blur intensities. 
% Visualizations for different $K_{blur}$ is provided in the supplementary.

\begin{figure}[t]
    \centering
    \setlength{\abovecaptionskip}{2mm}
    \includegraphics[width=\linewidth]{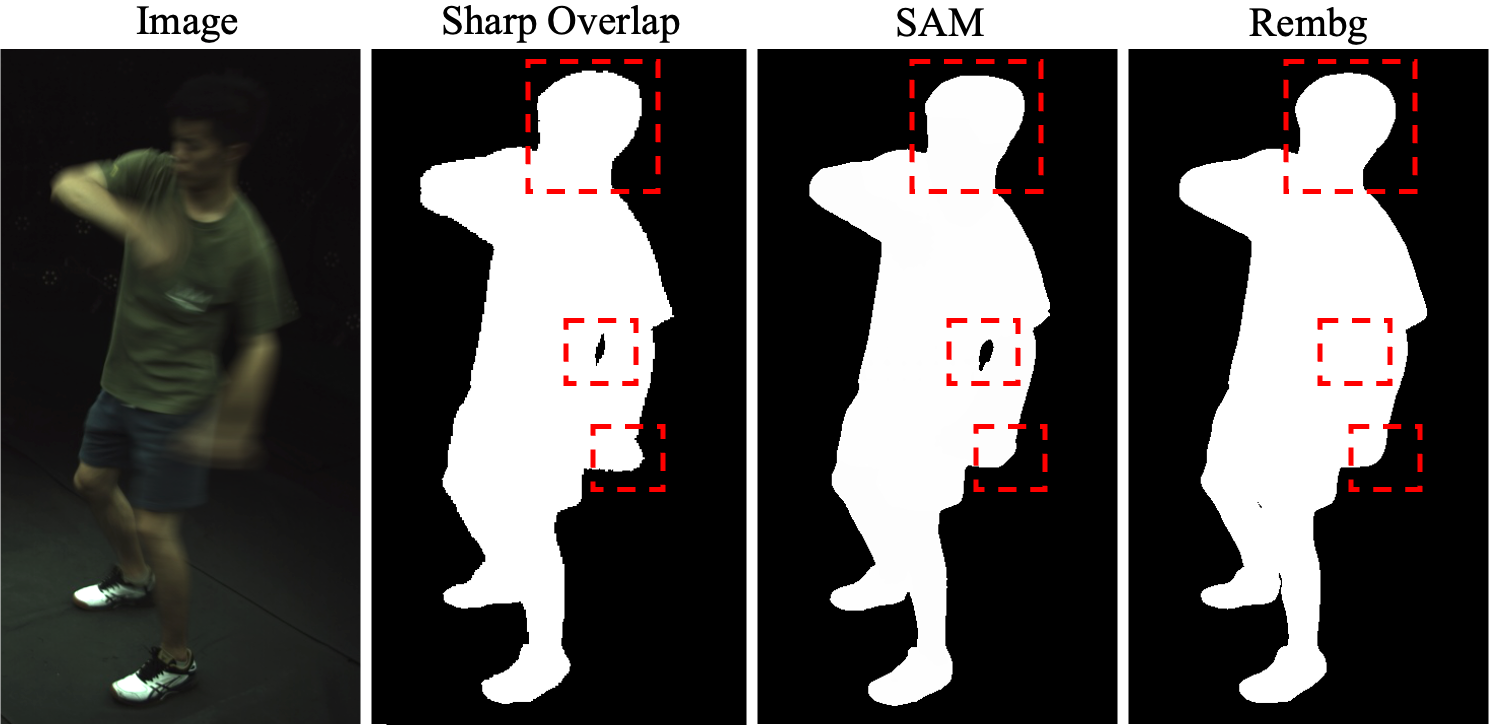}
    \caption{Visualization of different mask parsing results.}
    \label{fig:mask_different}
    % \vspace{-1mm}
\end{figure}

\begin{table}[t]
\centering
\setlength{\abovecaptionskip}{2mm}
\caption{Quantitative ablation results for different $K_{blur}$.}
\label{tab:ablation_Kblur}
\resizebox{\linewidth}{!}{%
\begin{tabular}{llccccc}
\hline
\multirow{2}{*}{$K_{blur}$} & \multirow{2}{*}{Metrics} & w/o & w/o pose & w/o & w/o & Ours \\
 & & interp. & deform & LBS opt. & shape opt. & (full model) \\
\hline
\multirow{3}{*}{$7$} & PSNR $\uparrow$ & 23.611 & 24.813 & \thirdscore{25.050} & \secondscore{25.153} & \bestscore{25.155} \\
 & SSIM $\uparrow$ & 0.7962 & 0.8127 & \thirdscore{0.8180} & \secondscore{0.8199} & \bestscore{0.8200} \\
 & LPIPS $\downarrow$ & 0.1710 & 0.1640 & \thirdscore{0.1563} & \secondscore{0.1561} & \bestscore{0.1557} \\
 \hline
\multirow{3}{*}{$9$} & PSNR $\uparrow$ & 22.800 & 24.239 & \thirdscore{24.596} & \bestscore{24.683} & \secondscore{24.680} \\
 & SSIM $\uparrow$ & 0.7811 & 0.8035 & \thirdscore{0.8109} & \bestscore{0.8127} & \secondscore{0.8126} \\
 & LPIPS $\downarrow$ & 0.1850 & 0.1734 & \secondscore{0.1638} & \thirdscore{0.1639} & \bestscore{0.1636}  \\
 \hline
\multirow{3}{*}{$11$} & PSNR $\uparrow$ & 22.350 & 23.754 & \thirdscore{24.282} & \secondscore{24.345} & \bestscore{24.353} \\
 & SSIM $\uparrow$ & 0.7695 & 0.7924 & \thirdscore{0.8022} & \secondscore{0.8037} & \bestscore{0.8039} \\
 & LPIPS $\downarrow$ & 0.1945 & 0.1852 & \thirdscore{0.1725} & \secondscore{0.1726} & \bestscore{0.1725} \\
\hline
\end{tabular}%
}
% \vspace{-1mm}
\end{table}

\begin{table}[t]
\centering
\setlength{\abovecaptionskip}{2mm}
\caption{Quantitative results with different mask estimations.}
\resizebox{0.65\linewidth}{!}{
\begin{tabular}{lccc}
\hline
Mask & PSNR $\uparrow$ & SSIM $\uparrow$ & LPIPS $\downarrow$ \\ \hline
sharp & 25.61 & .8309 & .1465 \\
SAM & 25.55 & .8290 & .1476 \\
rembg & 25.49 & .8281 & .1488 \\
\hline
\end{tabular}%
}
\label{tab:mask_robust}
% \vspace{-2mm}
\end{table}

\noindent \textbf{Different view numbers $N_{view}$.}
To assess robustness with respect to the view numbers, the proposed model is evaluated using varying numbers of training views $N_{view} = 2, 3, 4$. As shown in Tab.~\ref{tab:view_num}, the proposed method consistently outperforms baseline approaches with different $N_{view}$, demonstrating its strong generality. Furthermore, the ablation results under different $N_{view}$ in Tab.~\ref{tab:view_num_ablation} further confirm the effectiveness of each component.

\noindent \textbf{Mask estimation.}
Additional experiments are conducted to evaluate the model's robustness to inaccurate masks due to blurriness. Since the synthetic dataset is generated from sharp frames, an ``ideal" case is defined in which the masks are obtained by overlapping estimated masks from the corresponding sharp frames. Though such masks cannot be accessed in applications, by comparing to this idealized setting, the robustness of the model to mask estimation can be evaluated. Furthermore, alternative mask tools like \textit{rembg} are adapted. As summarized in Tab.~\ref{tab:mask_robust}, the proposed method performs comparably to the ideal case, and the variation across different mask models remains minimal. Fig.~\ref{fig:mask_different} visualizes different mask estimations.

\begin{figure}[t]
    \centering
    \setlength{\abovecaptionskip}{2mm}
    \includegraphics[width=\linewidth]{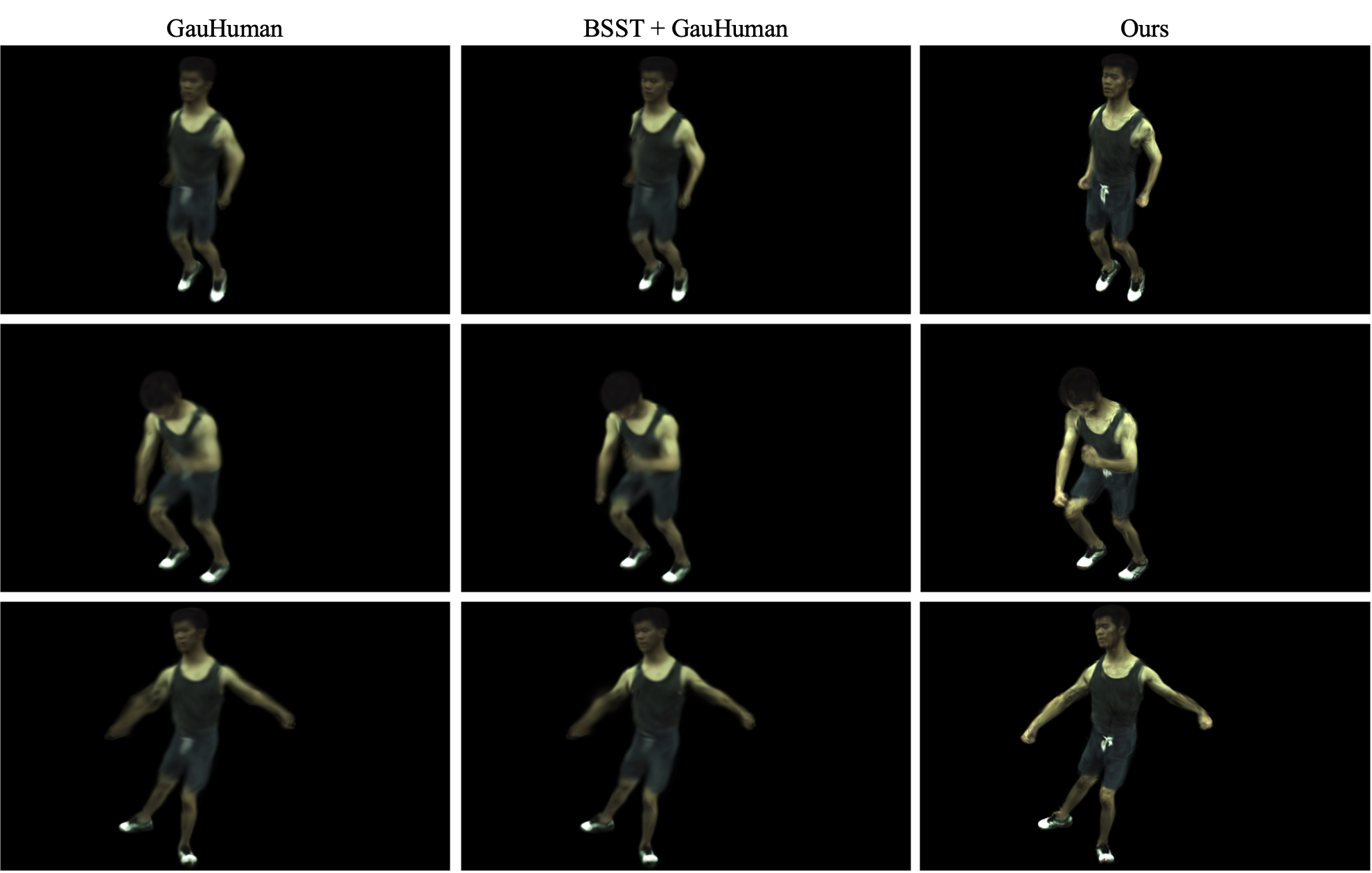}
    \caption{Novel pose animation results using out-of-domain motion from the AMASS~\cite{mahmood2019amass} dataset.}
    \label{fig:novel_pose}
    % \vspace{-2mm}
\end{figure}

\begin{table}[t]
\centering
\setlength{\abovecaptionskip}{2mm}
\caption{Quantitative comparisons with different $N_{view}$.}
\label{tab:view_num}
\resizebox{\linewidth}{!}{%
\begin{tabular}{llcccccc}
\hline
\multirow{2}{*}{$N_{view}$} & \multirow{2}{*}{Metrics} & \multirow{2}{*}{GauH} & ShiftNet & RVRT & VRT & BSST & \multirow{2}{*}{Ours} \\
 &  &  & + GauH & + GauH & + GauH & + GauH &  \\ \hline
\multirow{3}{*}{2} & PSNR $\uparrow$ & \secondscore{21.844} & 21.805 & \thirdscore{21.805} & 21.799 & 21.801 & \bestscore{23.331} \\
 & SSIM $\uparrow$ & 0.7501 & 0.7533 & \thirdscore{0.7533} & \secondscore{0.7534} & 0.7532 & \bestscore{0.7712} \\
 & LPIPS $\downarrow$ & 0.2506 & 0.2436 & \thirdscore{0.2431} & 0.2435 & \secondscore{0.2431} & \bestscore{0.1891} \\ \hline
\multirow{3}{*}{3} & PSNR $\uparrow$ & 22.514 & 22.580 & \secondscore{22.594} & 22.574 & \thirdscore{22.590} & \bestscore{24.979} \\
 & SSIM $\uparrow$ & 0.7545 & \secondscore{0.7584} & 0.7580 & 0.7580 & \thirdscore{0.7581} & \bestscore{0.8134} \\
 & LPIPS $\downarrow$ & 0.2317 & 0.2225 & \secondscore{0.2219} & \thirdscore{0.2225} & 0.2229 & \bestscore{0.1587} \\ \hline
\multirow{3}{*}{4} & PSNR $\uparrow$ & 23.080 & \secondscore{23.089} & 23.078 & 23.074 & \thirdscore{23.081} & \bestscore{25.546} \\
 & SSIM $\uparrow$ & 0.7660 & 0.7695 & \thirdscore{0.7697} & 0.7696 & \secondscore{0.7698} & \bestscore{0.8290} \\
 & LPIPS $\downarrow$ & 0.2277 & 0.2219 & \thirdscore{0.2218} & 0.2205 & \secondscore{0.2212} & \bestscore{0.1476} \\ \hline
\end{tabular}%
}
% \vspace{-2mm}
\end{table}

\noindent \textbf{Novel-pose evaluation.} 
Novel poses from the AMASS dataset~\cite{mahmood2019amass} are used to animate the reconstructed avatars for qualitative comparison. Fig.~\ref{fig:novel_pose} shows that the proposed model achieves the best result.

\subsection{Further Discussions}

\subsubsection{Demo captures from smartphones} 
\begin{figure}[t]
    \centering
    \setlength{\abovecaptionskip}{2mm}
    \includegraphics[width=\linewidth]{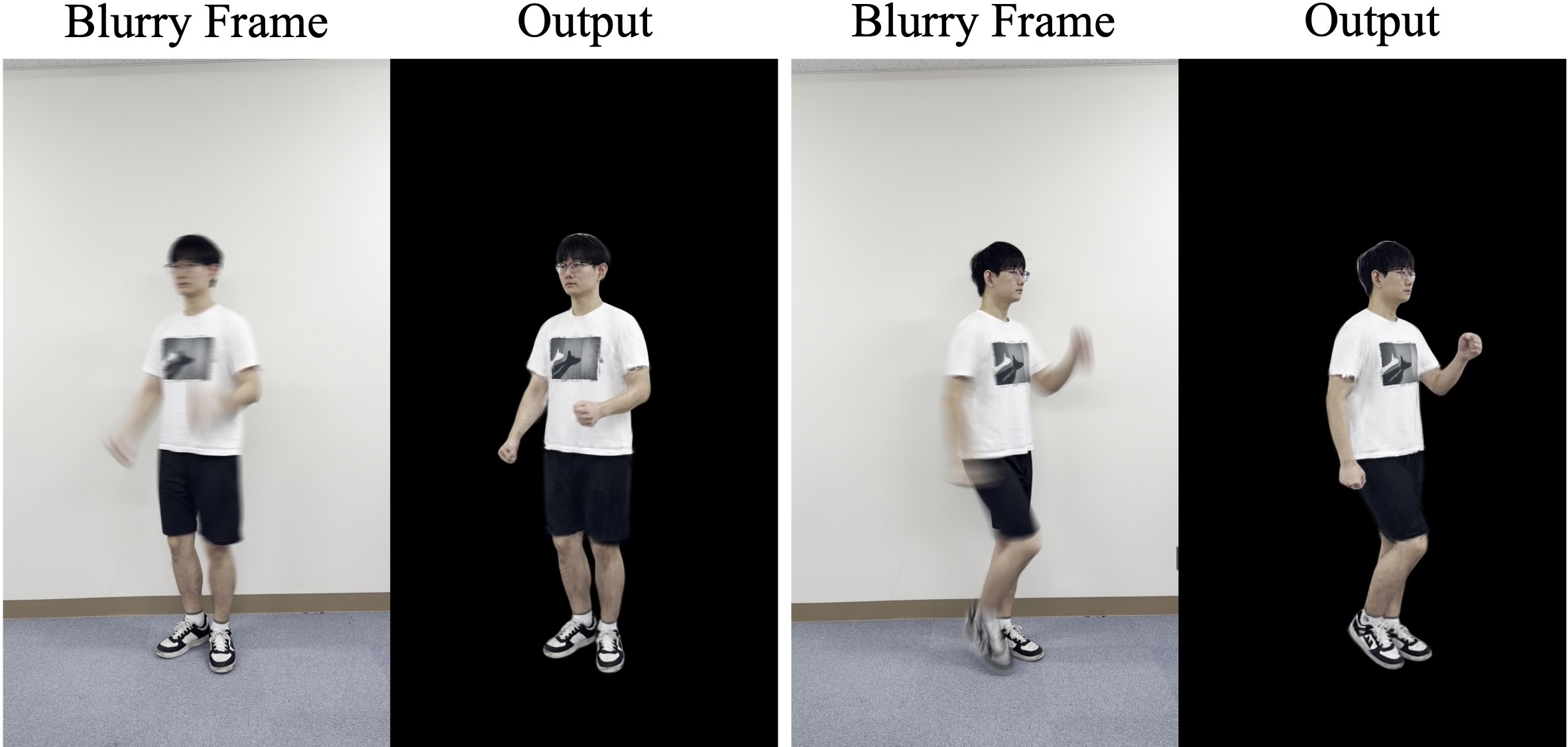}
    \caption{Qualitative results on captures from an iPhone 16 Pro.}
    \label{fig:iphone}
    % \vspace{-2mm}
\end{figure}

To further demonstrate the generality of the proposed method in real-world scenarios, monocular video is captured using an iPhone 16 Pro, and SMPL parameters are estimated with TRAM~\cite{wang2024tram}, an open-sourced human motion estimation model. Fig.~\ref{fig:iphone} confirm that the method generalizes well to real-world data and exhibits strong robustness to monocular SMPL estimations.

\subsubsection{Limitations} 
While the proposed method successfully estimates motion and generates sharp 3DGS avatars, it has certain limitations. For example, since the model is built upon 3DGS representation, it struggles to accurately recover the geometry of the person, \eg, the surface normal or the BRDF. In addition, real-world blur occurs as light integrates in the linear radiance space before the ISP process. Averaging non-linear sRGB values directly can lead to inaccuracies, particularly in high-contrast regions where linear summation is essential for physical fidelity. The sRGB space also make the method struggles on extreme situations like low-light degradations~\cite{niu2023visibility,niu2023nir,niu2023physics,guo2020zero,mildenhall2022nerf}. Future work aims to explore these aspects. Also, the proposed model relies on SMPL to represent sub-frame motion for 3D-aware deblurring. However, when the subject has handheld objects or wears loose garments, it fails to recover the motion of these accessories because SMPL does not have corresponding ``joints''. Future work could explore adaptive joints~\cite{zhan2025towards} with non-rigid modeling under blurriness to support more versatile avatar deblurring. Also, future works may leverage diffusion models~\cite{niu2024mofa,zhao2024cv,zhao2024stereocrafter,liu20243dgs,wu2025difix3d,niu2025anicrafter,chen2024mvsplat360} to further refine the avatar via generative prior information.

\begin{table}[t]
\centering
\setlength{\abovecaptionskip}{2mm}
\caption{Quantitative ablation results with different $N_{view}$.}
\label{tab:view_num_ablation}
\resizebox{\linewidth}{!}{%
\begin{tabular}{llccccc}
\hline
\multirow{2}{*}{$N_{view}$} & \multirow{2}{*}{Metrics} & w/o & w/o pose & w/o & w/o & \multirow{2}{*}{Ours} \\
 &  & interp. & deform & LBS opt. & shape opt. &  \\ \hline
\multirow{3}{*}{2} & PSNR $\uparrow$ & 22.661 & 23.310 & \thirdscore{23.321} & \bestscore{23.411} & \secondscore{23.331} \\
 & SSIM $\uparrow$ & 0.7584 & 0.7704 & \thirdscore{0.7718} & \bestscore{0.7737} & \secondscore{0.7712} \\
 & LPIPS $\downarrow$ & 0.1978 & 0.1919 & \thirdscore{0.1896} & \secondscore{0.1893} & \bestscore{0.1891} \\ \hline
\multirow{3}{*}{3} & PSNR $\uparrow$ & 23.927 & 24.808 & \thirdscore{24.899} & \secondscore{24.921} & \bestscore{24.979} \\
 & SSIM $\uparrow$ & 0.7973 & 0.8081 & \thirdscore{0.8115} & \secondscore{0.8129} & \bestscore{0.8134} \\
 & LPIPS $\downarrow$ & 0.1683 & 0.1656 & \thirdscore{0.1594} & \secondscore{0.1589} & \bestscore{0.1587} \\ \hline
\multirow{3}{*}{4} & PSNR $\uparrow$ & 24.009 & 25.301 & \thirdscore{25.394} & \secondscore{25.529} & \bestscore{25.546} \\
 & SSIM $\uparrow$ & 0.8053 & 0.8229 & \thirdscore{0.8261} & \secondscore{0.8284} & \bestscore{0.8290} \\
 & LPIPS $\downarrow$ & 0.1620 & 0.1545 & \thirdscore{0.1486} & \secondscore{0.1481} & \bestscore{0.1476} \\ \hline
\end{tabular}%
}
% \vspace{-2mm}
\end{table}

\section{Conclusion}
\label{sec:conclusion}

In this paper, a novel approach is presented for reconstructing sharp 3D human avatars from blurry video frames. By extending the traditional 2D motion blur formulation into a 3D-aware blur formation model, the proposed method jointly optimizes sub-frame motion representations while learning a canonical 3DGS human avatar model. To establish a benchmark for this task, both a synthetic dataset derived from the ZJU-MoCap and a real-captured dataset are provided. Extensive evaluations demonstrate that the proposed model consistently outperforms existing baselines.

\section*{Acknowledgment}

This research was supported in part by the Tateisi Science and Technology Foundation, JSPS KAKENHI Grant
Numbers 24KK0209 and 22H00529, and Advanced AI Talent Development to Lead the Next-Generation AI for Intelligent Society (BOOST NAIS) of The University of Tokyo.

{
    \small
    \bibliographystyle{ieeenat_fullname}
    \bibliography{main}

@String(CVPR= {IEEE Conf. Comput. Vis. Pattern Recog.})

@String(ICCV= {Int. Conf. Comput. Vis.})

@String(TOG= {ACM Trans. Graph.})

@String(CVPR  = {CVPR})

@String(ICCV  = {ICCV})

@String(TOG   = {ACM TOG})

@inproceedings{hyun2015generalized,
  title={Generalized video deblurring for dynamic scenes},
  author={Hyun Kim, Tae and Mu Lee, Kyoung},
  booktitle={Proceedings of the IEEE Conference on Computer Vision and Pattern Recognition},
  pages={5426--5434},
  year={2015}
}

@article{levin2006blind,
  title={Blind motion deblurring using image statistics},
  author={Levin, Anat},
  journal={Advances in neural information processing systems},
  volume={19},
  year={2006}
}

@inproceedings{ren2017video,
  title={Video deblurring via semantic segmentation and pixel-wise non-linear kernel},
  author={Ren, Wenqi and Pan, Jinshan and Cao, Xiaochun and Yang, Ming-Hsuan},
  booktitle={Proceedings of the IEEE International Conference on Computer Vision},
  pages={1077--1085},
  year={2017}
}

@inproceedings{wulff2014modeling,
  title={Modeling blurred video with layers},
  author={Wulff, Jonas and Black, Michael Julian},
  booktitle={European conference on computer vision},
  pages={236--252},
  year={2014},
  organization={Springer}
}

@inproceedings{nah2017deep,
  title={Deep multi-scale convolutional neural network for dynamic scene deblurring},
  author={Nah, Seungjun and Hyun Kim, Tae and Mu Lee, Kyoung},
  booktitle={Proceedings of the IEEE conference on computer vision and pattern recognition},
  pages={3883--3891},
  year={2017}
}

@inproceedings{su2017deep,
  title={Deep video deblurring for hand-held cameras},
  author={Su, Shuochen and Delbracio, Mauricio and Wang, Jue and Sapiro, Guillermo and Heidrich, Wolfgang and Wang, Oliver},
  booktitle={Proceedings of the IEEE conference on computer vision and pattern recognition},
  pages={1279--1288},
  year={2017}
}

@inproceedings{tao2018scale,
  title={Scale-recurrent network for deep image deblurring},
  author={Tao, Xin and Gao, Hongyun and Shen, Xiaoyong and Wang, Jue and Jia, Jiaya},
  booktitle={Proceedings of the IEEE conference on computer vision and pattern recognition},
  pages={8174--8182},
  year={2018}
}

@inproceedings{wang2019edvr,
  title={Edvr: Video restoration with enhanced deformable convolutional networks},
  author={Wang, Xintao and Chan, Kelvin CK and Yu, Ke and Dong, Chao and Change Loy, Chen},
  booktitle={Proceedings of the IEEE/CVF conference on computer vision and pattern recognition workshops},
  pages={0--0},
  year={2019}
}

@inproceedings{zamir2021multi,
  title={Multi-stage progressive image restoration},
  author={Zamir, Syed Waqas and Arora, Aditya and Khan, Salman and Hayat, Munawar and Khan, Fahad Shahbaz and Yang, Ming-Hsuan and Shao, Ling},
  booktitle={Proceedings of the IEEE/CVF conference on computer vision and pattern recognition},
  pages={14821--14831},
  year={2021}
}

@inproceedings{hyun2017online,
  title={Online video deblurring via dynamic temporal blending network},
  author={Hyun Kim, Tae and Mu Lee, Kyoung and Scholkopf, Bernhard and Hirsch, Michael},
  booktitle={Proceedings of the IEEE international conference on computer vision},
  pages={4038--4047},
  year={2017}
}

@inproceedings{nah2019recurrent,
  title={Recurrent neural networks with intra-frame iterations for video deblurring},
  author={Nah, Seungjun and Son, Sanghyun and Lee, Kyoung Mu},
  booktitle={Proceedings of the IEEE/CVF conference on computer vision and pattern recognition},
  pages={8102--8111},
  year={2019}
}

@inproceedings{wang2022efficient,
  title={Efficient video deblurring guided by motion magnitude},
  author={Wang, Yusheng and Lu, Yunfan and Gao, Ye and Wang, Lin and Zhong, Zhihang and Zheng, Yinqiang and Yamashita, Atsushi},
  booktitle={European Conference on Computer Vision},
  pages={413--429},
  year={2022},
  organization={Springer}
}

@inproceedings{zhong2020efficient,
  title={Efficient spatio-temporal recurrent neural network for video deblurring},
  author={Zhong, Zhihang and Gao, Ye and Zheng, Yinqiang and Zheng, Bo},
  booktitle={European conference on computer vision},
  pages={191--207},
  year={2020},
  organization={Springer}
}

@inproceedings{zhou2019spatio,
  title={Spatio-temporal filter adaptive network for video deblurring},
  author={Zhou, Shangchen and Zhang, Jiawei and Pan, Jinshan and Xie, Haozhe and Zuo, Wangmeng and Ren, Jimmy},
  booktitle={Proceedings of the IEEE/CVF international conference on computer vision},
  pages={2482--2491},
  year={2019}
}

@inproceedings{zhu2019deformable,
  title={Deformable convnets v2: More deformable, better results},
  author={Zhu, Xizhou and Hu, Han and Lin, Stephen and Dai, Jifeng},
  booktitle={Proceedings of the IEEE/CVF conference on computer vision and pattern recognition},
  pages={9308--9316},
  year={2019}
}

@inproceedings{dai2017deformable,
  title={Deformable convolutional networks},
  author={Dai, Jifeng and Qi, Haozhi and Xiong, Yuwen and Li, Yi and Zhang, Guodong and Hu, Han and Wei, Yichen},
  booktitle={Proceedings of the IEEE international conference on computer vision},
  pages={764--773},
  year={2017}
}

@inproceedings{pan2020cascaded,
  title={Cascaded deep video deblurring using temporal sharpness prior},
  author={Pan, Jinshan and Bai, Haoran and Tang, Jinhui},
  booktitle={Proceedings of the IEEE/CVF conference on computer vision and pattern recognition},
  pages={3043--3051},
  year={2020}
}

@article{son2021recurrent,
  title={Recurrent video deblurring with blur-invariant motion estimation and pixel volumes},
  author={Son, Hyeongseok and Lee, Junyong and Lee, Jonghyeop and Cho, Sunghyun and Lee, Seungyong},
  journal={ACM Transactions on Graphics (TOG)},
  volume={40},
  number={5},
  pages={1--18},
  year={2021},
  publisher={ACM New York, NY}
}

@inproceedings{kupyn2018deblurgan,
  title={Deblurgan: Blind motion deblurring using conditional adversarial networks},
  author={Kupyn, Orest and Budzan, Volodymyr and Mykhailych, Mykola and Mishkin, Dmytro and Matas, Ji{\v{r}}{\'\i}},
  booktitle={Proceedings of the IEEE conference on computer vision and pattern recognition},
  pages={8183--8192},
  year={2018}
}

@inproceedings{kupyn2019deblurgan,
  title={Deblurgan-v2: Deblurring (orders-of-magnitude) faster and better},
  author={Kupyn, Orest and Martyniuk, Tetiana and Wu, Junru and Wang, Zhangyang},
  booktitle={Proceedings of the IEEE/CVF international conference on computer vision},
  pages={8878--8887},
  year={2019}
}

@inproceedings{liang2021swinir,
  title={Swinir: Image restoration using swin transformer},
  author={Liang, Jingyun and Cao, Jiezhang and Sun, Guolei and Zhang, Kai and Van Gool, Luc and Timofte, Radu},
  booktitle={Proceedings of the IEEE/CVF international conference on computer vision},
  pages={1833--1844},
  year={2021}
}

@inproceedings{zamir2022restormer,
  title={Restormer: Efficient transformer for high-resolution image restoration},
  author={Zamir, Syed Waqas and Arora, Aditya and Khan, Salman and Hayat, Munawar and Khan, Fahad Shahbaz and Yang, Ming-Hsuan},
  booktitle={Proceedings of the IEEE/CVF conference on computer vision and pattern recognition},
  pages={5728--5739},
  year={2022}
}

@article{liang2022recurrent,
  title={Recurrent video restoration transformer with guided deformable attention},
  author={Liang, Jingyun and Fan, Yuchen and Xiang, Xiaoyu and Ranjan, Rakesh and Ilg, Eddy and Green, Simon and Cao, Jiezhang and Zhang, Kai and Timofte, Radu and Gool, Luc V},
  journal={Advances in Neural Information Processing Systems},
  volume={35},
  pages={378--393},
  year={2022}
}

@article{cao2022vdtr,
  title={VDTR: Video deblurring with transformer},
  author={Cao, Mingdeng and Fan, Yanbo and Zhang, Yong and Wang, Jue and Yang, Yujiu},
  journal={IEEE Transactions on Circuits and Systems for Video Technology},
  volume={33},
  number={1},
  pages={160--171},
  year={2022},
  publisher={IEEE}
}

@article{mildenhall2021nerf,
  title={Nerf: Representing scenes as neural radiance fields for view synthesis},
  author={Mildenhall, Ben and Srinivasan, Pratul P and Tancik, Matthew and Barron, Jonathan T and Ramamoorthi, Ravi and Ng, Ren},
  journal={Communications of the ACM},
  volume={65},
  number={1},
  pages={99--106},
  year={2021},
  publisher={ACM New York, NY, USA}
}

@inproceedings{noguchi2021neural,
  title={Neural articulated radiance field},
  author={Noguchi, Atsuhiro and Sun, Xiao and Lin, Stephen and Harada, Tatsuya},
  booktitle={Proceedings of the IEEE/CVF International Conference on Computer Vision},
  pages={5762--5772},
  year={2021}
}

@article{su2021nerf,
  title={A-nerf: Articulated neural radiance fields for learning human shape, appearance, and pose},
  author={Su, Shih-Yang and Yu, Frank and Zollh{\"o}fer, Michael and Rhodin, Helge},
  journal={Advances in neural information processing systems},
  volume={34},
  pages={12278--12291},
  year={2021}
}

@article{xu2021h,
  title={H-nerf: Neural radiance fields for rendering and temporal reconstruction of humans in motion},
  author={Xu, Hongyi and Alldieck, Thiemo and Sminchisescu, Cristian},
  journal={Advances in Neural Information Processing Systems},
  volume={34},
  pages={14955--14966},
  year={2021}
}

@inproceedings{guo2023vid2avatar,
  title={Vid2avatar: 3d avatar reconstruction from videos in the wild via self-supervised scene decomposition},
  author={Guo, Chen and Jiang, Tianjian and Chen, Xu and Song, Jie and Hilliges, Otmar},
  booktitle={Proceedings of the IEEE/CVF Conference on Computer Vision and Pattern Recognition},
  pages={12858--12868},
  year={2023}
}

@inproceedings{jiang2023instantavatar,
  title={Instantavatar: Learning avatars from monocular video in 60 seconds},
  author={Jiang, Tianjian and Chen, Xu and Song, Jie and Hilliges, Otmar},
  booktitle={Proceedings of the IEEE/CVF Conference on Computer Vision and Pattern Recognition},
  pages={16922--16932},
  year={2023}
}

@inproceedings{jiang2022neuman,
  title={Neuman: Neural human radiance field from a single video},
  author={Jiang, Wei and Yi, Kwang Moo and Samei, Golnoosh and Tuzel, Oncel and Ranjan, Anurag},
  booktitle={European Conference on Computer Vision},
  pages={402--418},
  year={2022},
  organization={Springer}
}

@inproceedings{li2022tava,
  title={Tava: Template-free animatable volumetric actors},
  author={Li, Ruilong and Tanke, Julian and Vo, Minh and Zollh{\"o}fer, Michael and Gall, J{\"u}rgen and Kanazawa, Angjoo and Lassner, Christoph},
  booktitle={European Conference on Computer Vision},
  pages={419--436},
  year={2022},
  organization={Springer}
}

@inproceedings{peng2021animatable,
  title={Animatable neural radiance fields for modeling dynamic human bodies},
  author={Peng, Sida and Dong, Junting and Wang, Qianqian and Zhang, Shangzhan and Shuai, Qing and Zhou, Xiaowei and Bao, Hujun},
  booktitle={Proceedings of the IEEE/CVF international conference on computer vision},
  pages={14314--14323},
  year={2021}
}

@misc{peng2022animatable,
      title={Animatable Implicit Neural Representations for Creating Realistic Avatars from Videos}, 
      author={Sida Peng and Zhen Xu and Junting Dong and Qianqian Wang and Shangzhan Zhang and Qing Shuai and Hujun Bao and Xiaowei Zhou},
      year={2023},
      eprint={2203.08133},
      archivePrefix={arXiv},
      primaryClass={cs.CV},
      url={https://arxiv.org/abs/2203.08133}, 
}

@inproceedings{wang2022arah,
  title={Arah: Animatable volume rendering of articulated human sdfs},
  author={Wang, Shaofei and Schwarz, Katja and Geiger, Andreas and Tang, Siyu},
  booktitle={European conference on computer vision},
  pages={1--19},
  year={2022},
  organization={Springer}
}

@inproceedings{weng2022humannerf,
  title={Humannerf: Free-viewpoint rendering of moving people from monocular video},
  author={Weng, Chung-Yi and Curless, Brian and Srinivasan, Pratul P and Barron, Jonathan T and Kemelmacher-Shlizerman, Ira},
  booktitle={Proceedings of the IEEE/CVF conference on computer vision and pattern Recognition},
  pages={16210--16220},
  year={2022}
}

@inproceedings{yu2023monohuman,
  title={Monohuman: Animatable human neural field from monocular video},
  author={Yu, Zhengming and Cheng, Wei and Liu, Xian and Wu, Wayne and Lin, Kwan-Yee},
  booktitle={Proceedings of the IEEE/CVF Conference on Computer Vision and Pattern Recognition},
  pages={16943--16953},
  year={2023}
}

@article{kerbl20233d,
  title={3d gaussian splatting for real-time radiance field rendering.},
  author={Kerbl, Bernhard and Kopanas, Georgios and Leimk{\"u}hler, Thomas and Drettakis, George},
  journal={ACM Trans. Graph.},
  volume={42},
  number={4},
  pages={139--1},
  year={2023}
}

@inproceedings{zielonka2025drivable,
  title={Drivable 3d gaussian avatars},
  author={Zielonka, Wojciech and Bagautdinov, Timur and Saito, Shunsuke and Zollh{\"o}fer, Michael and Thies, Justus and Romero, Javier},
  booktitle={2025 International Conference on 3D Vision (3DV)},
  pages={979--990},
  year={2025},
  organization={IEEE}
}

@inproceedings{li2024animatable,
  title={Animatable gaussians: Learning pose-dependent gaussian maps for high-fidelity human avatar modeling},
  author={Li, Zhe and Zheng, Zerong and Wang, Lizhen and Liu, Yebin},
  booktitle={Proceedings of the IEEE/CVF conference on computer vision and pattern recognition},
  pages={19711--19722},
  year={2024}
}

@article{jena2023splatarmor,
  title={Splatarmor: Articulated gaussian splatting for animatable humans from monocular rgb videos},
  author={Jena, Rohit and Iyer, Ganesh Subramanian and Choudhary, Siddharth and Smith, Brandon and Chaudhari, Pratik and Gee, James},
  journal={arXiv preprint arXiv:2311.10812},
  year={2023}
}

@inproceedings{moreau2024human,
  title={Human gaussian splatting: Real-time rendering of animatable avatars},
  author={Moreau, Arthur and Song, Jifei and Dhamo, Helisa and Shaw, Richard and Zhou, Yiren and P{\'e}rez-Pellitero, Eduardo},
  booktitle={Proceedings of the IEEE/CVF conference on computer vision and pattern recognition},
  pages={788--798},
  year={2024}
}

@article{ye2023animatable,
  title={Animatable 3d gaussians for high-fidelity synthesis of human motions},
  author={Ye, Keyang and Shao, Tianjia and Zhou, Kun},
  journal={arXiv preprint arXiv:2311.13404},
  year={2023}
}

@inproceedings{kocabas2024hugs,
  title={Hugs: Human gaussian splats},
  author={Kocabas, Muhammed and Chang, Jen-Hao Rick and Gabriel, James and Tuzel, Oncel and Ranjan, Anurag},
  booktitle={Proceedings of the IEEE/CVF conference on computer vision and pattern recognition},
  pages={505--515},
  year={2024}
}

@inproceedings{lei2024gart,
  title={Gart: Gaussian articulated template models},
  author={Lei, Jiahui and Wang, Yufu and Pavlakos, Georgios and Liu, Lingjie and Daniilidis, Kostas},
  booktitle={Proceedings of the IEEE/CVF conference on computer vision and pattern recognition},
  pages={19876--19887},
  year={2024}
}

@inproceedings{liu2024animatable,
  title={Animatable 3d gaussian: Fast and high-quality reconstruction of multiple human avatars},
  author={Liu, Yang and Huang, Xiang and Qin, Minghan and Lin, Qinwei and Wang, Haoqian},
  booktitle={Proceedings of the 32nd ACM International Conference on Multimedia},
  pages={1120--1129},
  year={2024}
}

@inproceedings{hu2024gauhuman,
  title={Gauhuman: Articulated gaussian splatting from monocular human videos},
  author={Hu, Shoukang and Hu, Tao and Liu, Ziwei},
  booktitle={Proceedings of the IEEE/CVF conference on computer vision and pattern recognition},
  pages={20418--20431},
  year={2024}
}

@inproceedings{qian20243dgs,
  title={3dgs-avatar: Animatable avatars via deformable 3d gaussian splatting},
  author={Qian, Zhiyin and Wang, Shaofei and Mihajlovic, Marko and Geiger, Andreas and Tang, Siyu},
  booktitle={Proceedings of the IEEE/CVF conference on computer vision and pattern recognition},
  pages={5020--5030},
  year={2024}
}

@inproceedings{hu2024gaussianavatar,
  title={Gaussianavatar: Towards realistic human avatar modeling from a single video via animatable 3d gaussians},
  author={Hu, Liangxiao and Zhang, Hongwen and Zhang, Yuxiang and Zhou, Boyao and Liu, Boning and Zhang, Shengping and Nie, Liqiang},
  booktitle={Proceedings of the IEEE/CVF conference on computer vision and pattern recognition},
  pages={634--644},
  year={2024}
}

@inproceedings{zhong2023blur,
  title={Blur interpolation transformer for real-world motion from blur},
  author={Zhong, Zhihang and Cao, Mingdeng and Ji, Xiang and Zheng, Yinqiang and Sato, Imari},
  booktitle={Proceedings of the IEEE/CVF Conference on Computer Vision and Pattern Recognition},
  pages={5713--5723},
  year={2023}
}

@inproceedings{cao2022towards,
  title={Towards real-world video deblurring by exploring blur formation process},
  author={Cao, Mingdeng and Zhong, Zhihang and Fan, Yanbo and Wang, Jiahao and Zhang, Yong and Wang, Jue and Yang, Yujiu and Zheng, Yinqiang},
  booktitle={European Conference on Computer Vision},
  pages={327--343},
  year={2022},
  organization={Springer}
}

@inproceedings{zhong2022animation,
  title={Animation from blur: Multi-modal blur decomposition with motion guidance},
  author={Zhong, Zhihang and Sun, Xiao and Wu, Zhirong and Zheng, Yinqiang and Lin, Stephen and Sato, Imari},
  booktitle={European Conference on Computer Vision},
  pages={599--615},
  year={2022},
  organization={Springer}
}

@inproceedings{ma2022deblur,
  title={Deblur-nerf: Neural radiance fields from blurry images},
  author={Ma, Li and Li, Xiaoyu and Liao, Jing and Zhang, Qi and Wang, Xuan and Wang, Jue and Sander, Pedro V},
  booktitle={Proceedings of the IEEE/CVF conference on computer vision and pattern recognition},
  pages={12861--12870},
  year={2022}
}

@inproceedings{wang2023bad,
  title={Bad-nerf: Bundle adjusted deblur neural radiance fields},
  author={Wang, Peng and Zhao, Lingzhe and Ma, Ruijie and Liu, Peidong},
  booktitle={Proceedings of the IEEE/CVF Conference on Computer Vision and Pattern Recognition},
  pages={4170--4179},
  year={2023}
}

@inproceedings{lee2023dp,
  title={Dp-nerf: Deblurred neural radiance field with physical scene priors},
  author={Lee, Dogyoon and Lee, Minhyeok and Shin, Chajin and Lee, Sangyoun},
  booktitle={Proceedings of the IEEE/CVF Conference on Computer Vision and Pattern Recognition},
  pages={12386--12396},
  year={2023}
}

@inproceedings{zhao2024bad,
  title={Bad-gaussians: Bundle adjusted deblur gaussian splatting},
  author={Zhao, Lingzhe and Wang, Peng and Liu, Peidong},
  booktitle={European Conference on Computer Vision},
  pages={233--250},
  year={2024},
  organization={Springer}
}

@inproceedings{sun2024dyblurf,
  title={Dyblurf: Dynamic neural radiance fields from blurry monocular video},
  author={Sun, Huiqiang and Li, Xingyi and Shen, Liao and Ye, Xinyi and Xian, Ke and Cao, Zhiguo},
  booktitle={Proceedings of the IEEE/CVF Conference on Computer Vision and Pattern Recognition},
  pages={7517--7527},
  year={2024}
}

@article{bui2025moblurf,
  title={Moblurf: Motion deblurring neural radiance fields for blurry monocular video},
  author={Bui, Minh-Quan Viet and Park, Jongmin and Oh, Jihyong and Kim, Munchurl},
  journal={IEEE Transactions on Pattern Analysis and Machine Intelligence},
  year={2025},
  publisher={IEEE}
}

@inproceedings{stearns2024dynamic,
  title={Dynamic gaussian marbles for novel view synthesis of casual monocular videos},
  author={Stearns, Colton and Harley, Adam and Uy, Mikaela and Dubost, Florian and Tombari, Federico and Wetzstein, Gordon and Guibas, Leonidas},
  booktitle={SIGGRAPH Asia 2024 Conference Papers},
  pages={1--11},
  year={2024}
}

@inproceedings{peng2021neural,
  title={Neural body: Implicit neural representations with structured latent codes for novel view synthesis of dynamic humans},
  author={Peng, Sida and Zhang, Yuanqing and Xu, Yinghao and Wang, Qianqian and Shuai, Qing and Bao, Hujun and Zhou, Xiaowei},
  booktitle={Proceedings of the IEEE/CVF conference on computer vision and pattern recognition},
  pages={9054--9063},
  year={2021}
}

@inproceedings{cha2023generating,
  title={Generating texture for 3d human avatar from a single image using sampling and refinement networks},
  author={Cha, Sihun and Seo, Kwanggyoon and Ashtari, Amirsaman and Noh, Junyong},
  booktitle={Computer graphics forum},
  volume={42},
  number={2},
  pages={385--396},
  year={2023},
  organization={Wiley Online Library}
}

@inproceedings{huang2024tech,
  title={Tech: Text-guided reconstruction of lifelike clothed humans},
  author={Huang, Yangyi and Yi, Hongwei and Xiu, Yuliang and Liao, Tingting and Tang, Jiaxiang and Cai, Deng and Thies, Justus},
  booktitle={2024 International Conference on 3D Vision (3DV)},
  pages={1531--1542},
  year={2024},
  organization={IEEE}
}

@inproceedings{liao2023high,
  title={High-fidelity clothed avatar reconstruction from a single image},
  author={Liao, Tingting and Zhang, Xiaomei and Xiu, Yuliang and Yi, Hongwei and Liu, Xudong and Qi, Guo-Jun and Zhang, Yong and Wang, Xuan and Zhu, Xiangyu and Lei, Zhen},
  booktitle={Proceedings of the IEEE/CVF conference on computer vision and pattern recognition},
  pages={8662--8672},
  year={2023}
}

@inproceedings{zhang2024humanref,
  title={Humanref: Single image to 3d human generation via reference-guided diffusion},
  author={Zhang, Jingbo and Li, Xiaoyu and Zhang, Qi and Cao, Yanpei and Shan, Ying and Liao, Jing},
  booktitle={Proceedings of the IEEE/CVF Conference on Computer Vision and Pattern Recognition},
  pages={1844--1854},
  year={2024}
}

@inproceedings{pan2023deep,
  title={Deep discriminative spatial and temporal network for efficient video deblurring},
  author={Pan, Jinshan and Xu, Boming and Dong, Jiangxin and Ge, Jianjun and Tang, Jinhui},
  booktitle={Proceedings of the IEEE/CVF Conference on Computer Vision and Pattern Recognition},
  pages={22191--22200},
  year={2023}
}

@article{liang2024vrt,
  title={Vrt: A video restoration transformer},
  author={Liang, Jingyun and Cao, Jiezhang and Fan, Yuchen and Zhang, Kai and Ranjan, Rakesh and Li, Yawei and Timofte, Radu and Van Gool, Luc},
  journal={IEEE Transactions on Image Processing},
  volume={33},
  pages={2171--2182},
  year={2024},
  publisher={IEEE}
}

@incollection{loper2023smpl,
  title={SMPL: A skinned multi-person linear model},
  author={Loper, Matthew and Mahmood, Naureen and Romero, Javier and Pons-Moll, Gerard and Black, Michael J},
  booktitle={Seminal Graphics Papers: Pushing the Boundaries, Volume 2},
  pages={851--866},
  year={2023}
}

@inproceedings{qin1998general,
  title={General matrix representations for B-splines},
  author={Qin, Kaihuai},
  booktitle={Proceedings Pacific Graphics' 98. Sixth Pacific Conference on Computer Graphics and Applications (Cat. No. 98EX208)},
  pages={37--43},
  year={1998},
  organization={IEEE}
}

@book{farin2002curves,
  title={Curves and surfaces for CAGD: a practical guide},
  author={Farin, Gerald E},
  year={2002},
  publisher={Morgan Kaufmann}
}

@article{unser2002b,
  title={B-spline signal processing. I. Theory},
  author={Unser, Michael and Aldroubi, Akram and Eden, Murray},
  journal={IEEE transactions on signal processing},
  volume={41},
  number={2},
  pages={821--833},
  year={2002},
  publisher={IEEE}
}

@article{kingma2014adam,
  title={Adam: A method for stochastic optimization},
  author={Kingma, Diederik P and Ba, Jimmy},
  journal={arXiv preprint arXiv:1412.6980},
  year={2014}
}

@inproceedings{geng2023learning,
  title={Learning neural volumetric representations of dynamic humans in minutes},
  author={Geng, Chen and Peng, Sida and Xu, Zhen and Bao, Hujun and Zhou, Xiaowei},
  booktitle={Proceedings of the IEEE/CVF Conference on Computer Vision and Pattern Recognition},
  pages={8759--8770},
  year={2023}
}

@inproceedings{huang2022real,
  title={Real-time intermediate flow estimation for video frame interpolation},
  author={Huang, Zhewei and Zhang, Tianyuan and Heng, Wen and Shi, Boxin and Zhou, Shuchang},
  booktitle={European conference on computer vision},
  pages={624--642},
  year={2022},
  organization={Springer}
}

@inproceedings{li2023simple,
  title={A simple baseline for video restoration with grouped spatial-temporal shift},
  author={Li, Dasong and Shi, Xiaoyu and Zhang, Yi and Cheung, Ka Chun and See, Simon and Wang, Xiaogang and Qin, Hongwei and Li, Hongsheng},
  booktitle={Proceedings of the IEEE/CVF Conference on Computer Vision and Pattern Recognition},
  pages={9822--9832},
  year={2023}
}

@inproceedings{zhang2024blur,
  title={Blur-aware spatio-temporal sparse transformer for video deblurring},
  author={Zhang, Huicong and Xie, Haozhe and Yao, Hongxun},
  booktitle={Proceedings of the IEEE/CVF Conference on Computer Vision and Pattern Recognition},
  pages={2673--2681},
  year={2024}
}

@article{li2023usb,
  title={Usb-nerf: Unrolling shutter bundle adjusted neural radiance fields},
  author={Li, Moyang and Wang, Peng and Zhao, Lingzhe and Liao, Bangyan and Liu, Peidong},
  journal={arXiv preprint arXiv:2310.02687},
  year={2023}
}

@inproceedings{kirillov2023segment,
  title={Segment anything},
  author={Kirillov, Alexander and Mintun, Eric and Ravi, Nikhila and Mao, Hanzi and Rolland, Chloe and Gustafson, Laura and Xiao, Tete and Whitehead, Spencer and Berg, Alexander C and Lo, Wan-Yen and others},
  booktitle={Proceedings of the IEEE/CVF international conference on computer vision},
  pages={4015--4026},
  year={2023}
}

@inproceedings{wang2024tram,
  title={Tram: Global trajectory and motion of 3d humans from in-the-wild videos},
  author={Wang, Yufu and Wang, Ziyun and Liu, Lingjie and Daniilidis, Kostas},
  booktitle={European Conference on Computer Vision},
  pages={467--487},
  year={2024},
  organization={Springer}
}

@inproceedings{lu2025bard,
  title={Bard-gs: Blur-aware reconstruction of dynamic scenes via gaussian splatting},
  author={Lu, Yiren and Zhou, Yunlai and Liu, Disheng and Liang, Tuo and Yin, Yu},
  booktitle={Proceedings of the Computer Vision and Pattern Recognition Conference},
  pages={16532--16542},
  year={2025}
}

@inproceedings{kim2024motion,
  title={Motion-oriented compositional neural radiance fields for monocular dynamic human modeling},
  author={Kim, Jaehyeok and Wee, Dongyoon and Xu, Dan},
  booktitle={European Conference on Computer Vision},
  pages={476--493},
  year={2024},
  organization={Springer}
}

@article{yin2025event,
  title={Event-guided 3D Gaussian Splatting for Dynamic Human and Scene Reconstruction},
  author={Yin, Xiaoting and Shi, Hao and Yang, Kailun and Zhai, Jiajun and Guo, Shangwei and Wang, Lin and Wang, Kaiwei},
  journal={arXiv preprint arXiv:2509.18566},
  year={2025}
}

@inproceedings{mahmood2019amass,
  title={AMASS: Archive of motion capture as surface shapes},
  author={Mahmood, Naureen and Ghorbani, Nima and Troje, Nikolaus F and Pons-Moll, Gerard and Black, Michael J},
  booktitle={Proceedings of the IEEE/CVF international conference on computer vision},
  pages={5442--5451},
  year={2019}
}

@inproceedings{zhan2025towards,
  title={Towards Explicit Exoskeleton for the Reconstruction of Complicated 3D Human Avatars},
  author={Zhan, Yifan and Zhu, Qingtian and Niu, Muyao and Ma, Mingze and Zhao, Jiancheng and Zhong, Zhihang and Sun, Xiao and Qiao, Yu and Zheng, Yinqiang},
  booktitle={Proceedings of the IEEE/CVF International Conference on Computer Vision},
  pages={14259--14269},
  year={2025}
}

@inproceedings{xu2025sequential,
  title={Sequential Gaussian Avatars with Hierarchical Motion Context},
  author={Xu, Wangze and Zhan, Yifan and Zhong, Zhihang and Sun, Xiao},
  booktitle={Proceedings of the IEEE/CVF International Conference on Computer Vision},
  pages={13592--13603},
  year={2025}
}

@inproceedings{chen2024within,
  title={Within the dynamic context: Inertia-aware 3d human modeling with pose sequence},
  author={Chen, Yutong and Zhan, Yifan and Zhong, Zhihang and Wang, Wei and Sun, Xiao and Qiao, Yu and Zheng, Yinqiang},
  booktitle={European Conference on Computer Vision},
  pages={491--508},
  year={2024},
  organization={Springer}
}

@article{niu2025anicrafter,
  title={Anicrafter: Customizing realistic human-centric animation via avatar-background conditioning in video diffusion models},
  author={Niu, Muyao and Cao, Mingdeng and Zhan, Yifan and Zhu, Qingtian and Ma, Mingze and Zhao, Jiancheng and Zeng, Yanhong and Zhong, Zhihang and Sun, Xiao and Zheng, Yinqiang},
  journal={arXiv preprint arXiv:2505.20255},
  year={2025}
}

@article{zhong2023real,
  title={Real-world video deblurring: A benchmark dataset and an efficient recurrent neural network},
  author={Zhong, Zhihang and Gao, Ye and Zheng, Yinqiang and Zheng, Bo and Sato, Imari},
  journal={International Journal of Computer Vision},
  volume={131},
  number={1},
  pages={284--301},
  year={2023},
  publisher={Springer}
}

@misc{gao2025proxy,
      title={Proxy-GS: Unified Occlusion Priors for Training and Inference in Structured 3D Gaussian Splatting}, 
      author={Yuanyuan Gao and Yuning Gong and Yifei Liu and Li Jingfeng and Dingwen Zhang and Yanci Zhang and Dan Xu and Xiao Sun and Zhihang Zhong},
      year={2026},
      eprint={2509.24421},
      archivePrefix={arXiv},
      primaryClass={cs.CV},
      url={https://arxiv.org/abs/2509.24421}, 
}

@inproceedings{liu2025maskgaussian,
  title={Maskgaussian: Adaptive 3d gaussian representation from probabilistic masks},
  author={Liu, Yifei and Zhong, Zhihang and Zhan, Yifan and Xu, Sheng and Sun, Xiao},
  booktitle={Proceedings of the Computer Vision and Pattern Recognition Conference},
  pages={681--690},
  year={2025}
}

@inproceedings{gao2025citygs,
  title={Citygs-x: A scalable architecture for efficient and geometrically accurate large-scale scene reconstruction},
  author={Gao, Yuanyuan and Li, Hao and Chen, Jiaqi and Zou, Zhengyu and Zhong, Zhihang and Zhang, Dingwen and Sun, Xiao and Han, Junwei},
  booktitle={Proceedings of the IEEE/CVF International Conference on Computer Vision},
  pages={27187--27196},
  year={2025}
}

@article{zhan2025r3,
  title={R3-avatar: Record and retrieve temporal codebook for reconstructing photorealistic human avatars},
  author={Zhan, Yifan and Xu, Wangze and Zhu, Qingtian and Niu, Muyao and Ma, Mingze and Liu, Yifei and Zhong, Zhihang and Sun, Xiao and Zheng, Yinqiang},
  journal={arXiv preprint arXiv:2503.12751},
  year={2025}
}

@inproceedings{zhong2021towards,
  title={Towards rolling shutter correction and deblurring in dynamic scenes},
  author={Zhong, Zhihang and Zheng, Yinqiang and Sato, Imari},
  booktitle={Proceedings of the IEEE/CVF Conference on Computer Vision and Pattern Recognition},
  pages={9219--9228},
  year={2021}
}

@inproceedings{niu2024rs,
  title={Rs-nerf: Neural radiance fields from rolling shutter images},
  author={Niu, Muyao and Chen, Tong and Zhan, Yifan and Li, Zhuoxiao and Ji, Xiang and Zheng, Yinqiang},
  booktitle={European Conference on Computer Vision},
  pages={163--180},
  year={2024},
  organization={Springer}
}

@inproceedings{niu2024mofa,
  title={Mofa-video: Controllable image animation via generative motion field adaptions in frozen image-to-video diffusion model},
  author={Niu, Muyao and Cun, Xiaodong and Wang, Xintao and Zhang, Yong and Shan, Ying and Zheng, Yinqiang},
  booktitle={European conference on computer vision},
  pages={111--128},
  year={2024},
  organization={Springer}
}

@inproceedings{niu2023visibility,
  title={Visibility constrained wide-band illumination spectrum design for seeing-in-the-dark},
  author={Niu, Muyao and Li, Zhuoxiao and Zhong, Zhihang and Zheng, Yinqiang},
  booktitle={2023 IEEE/CVF Conference on Computer Vision and Pattern Recognition (CVPR)},
  pages={13976--13985},
  year={2023},
  organization={IEEE}
}

@inproceedings{niu2023nir,
  title={NIR-assisted video enhancement via unpaired 24-hour data},
  author={Niu, Muyao and Zhong, Zhihang and Zheng, Yinqiang},
  booktitle={2023 IEEE/CVF International Conference on Computer Vision (ICCV)},
  pages={10744--10754},
  year={2023},
  organization={IEEE}
}

@article{zhao2024cv,
  title={Cv-vae: A compatible video vae for latent generative video models},
  author={Zhao, Sijie and Zhang, Yong and Cun, Xiaodong and Yang, Shaoshu and Niu, Muyao and Li, Xiaoyu and Hu, Wenbo and Shan, Ying},
  journal={Advances in Neural Information Processing Systems},
  volume={37},
  pages={12847--12871},
  year={2024}
}

@article{zhao2024stereocrafter,
  title={Stereocrafter: Diffusion-based generation of long and high-fidelity stereoscopic 3d from monocular videos},
  author={Zhao, Sijie and Hu, Wenbo and Cun, Xiaodong and Zhang, Yong and Li, Xiaoyu and Kong, Zhe and Gao, Xiangjun and Niu, Muyao and Shan, Ying},
  journal={arXiv preprint arXiv:2409.07447},
  year={2024}
}

@article{liu20243dgs,
  title={3dgs-enhancer: Enhancing unbounded 3d gaussian splatting with view-consistent 2d diffusion priors},
  author={Liu, Xi and Zhou, Chaoyi and Huang, Siyu},
  journal={Advances in Neural Information Processing Systems},
  volume={37},
  pages={133305--133327},
  year={2024}
}

@inproceedings{wu2025difix3d,
  title={Difix3d+: Improving 3d reconstructions with single-step diffusion models},
  author={Wu, Jay Zhangjie and Zhang, Yuxuan and Turki, Haithem and Ren, Xuanchi and Gao, Jun and Shou, Mike Zheng and Fidler, Sanja and Gojcic, Zan and Ling, Huan},
  booktitle={2025 IEEE/CVF Conference on Computer Vision and Pattern Recognition (CVPR)},
  pages={26024--26035},
  year={2025},
  organization={IEEE}
}

@article{chen2024mvsplat360,
  title={Mvsplat360: Feed-forward 360 scene synthesis from sparse views},
  author={Chen, Yuedong and Zheng, Chuanxia and Xu, Haofei and Zhuang, Bohan and Vedaldi, Andrea and Cham, Tat-Jen and Cai, Jianfei},
  journal={Advances in Neural Information Processing Systems},
  volume={37},
  pages={107064--107086},
  year={2024}
}

@inproceedings{niu2023physics,
  title={Physics-based adversarial attack on near-infrared human detector for nighttime surveillance camera systems},
  author={Niu, Muyao and Li, Zhuoxiao and Zhan, Yifan and Nguyen, Huy H and Echizen, Isao and Zheng, Yinqiang},
  booktitle={Proceedings of the 31st ACM International Conference on Multimedia},
  pages={8799--8807},
  year={2023}
}

@inproceedings{guo2020zero,
  title={Zero-reference deep curve estimation for low-light image enhancement},
  author={Guo, Chunle and Li, Chongyi and Guo, Jichang and Loy, Chen Change and Hou, Junhui and Kwong, Sam and Cong, Runmin},
  booktitle={2020 IEEE/CVF Conference on Computer Vision and Pattern Recognition (CVPR)},
  pages={1777--1786},
  year={2020},
  organization={Ieee}
}

@inproceedings{mildenhall2022nerf,
  title={Nerf in the dark: High dynamic range view synthesis from noisy raw images},
  author={Mildenhall, Ben and Hedman, Peter and Martin-Brualla, Ricardo and Srinivasan, Pratul P and Barron, Jonathan T},
  booktitle={2022 IEEE/CVF conference on computer vision and pattern recognition (CVPR)},
  pages={16169--16178},
  year={2022},
  organization={IEEE}
}
}

\end{document}